%% file: main-arxiv.tex
\documentclass{article}
\usepackage{graphicx} 
\usepackage{amssymb} 
\usepackage{xcolor, amsmath, amsthm}
\usepackage[english]{babel}
\usepackage[authoryear,semicolon]{natbib}
\usepackage{hyperref}
\usepackage{cleveref}

\usepackage{enumitem}
\usepackage{svg}
\usepackage{xcolor}
\usepackage{import}
\usepackage{tikz}

\usetikzlibrary{automata,positioning, calc, shapes, arrows, fit, backgrounds, shapes.geometric, decorations.markings}

\title{Temporal logics and formal synthesis for robot planning and control}
\author{Jana Tumova\\tumova@kth.se
\and Joris Verhagen\\ jorisv@kth.se
\and Matti Vahs \\ vahs@kth.se \and \\{
KTH Royal Institute of Technology}}
\date{}

\newtheorem{definition}{Definition}
\newtheorem{problem}{Problem}
\newtheorem{example}{Example}
\newtheorem{remark}{Remark}

\newcommand{\U}{\, \mathcal U \,}
\newcommand{\UI}{\, \mathcal U_I \,}
\newcommand{\X}{\mathcal X \,}
\newcommand{\F}{\mathcal F \,}
\newcommand{\G}{\mathcal G \,}
\newcommand{\FI}{\mathcal F_I \,}
\newcommand{\GI}{\mathcal G_I \,}
\newcommand{\AP}{\mathit{AP}}

\newcommand{\T}{\mathcal T}
\newcommand{\A}{\mathcal A}
\newcommand{\Prod}{\mathcal P}

\newcommand{\sinit}{s_\mathit{init}}
\newcommand{\qinit}{q_\mathit{init}}

\begin{document}

\maketitle

\begin{abstract}
    As robots move from controlled environments into real-world settings, it becomes increasingly crucial to ensure that they perform as expected. A key step toward that goal is a rigorous specification of the desired robot behavior, capturing intricate temporal, spatial, and logical requirements. Complementing this, plan and control synthesis methods are needed to fulfill these specifications with provable guarantees. This manuscript presents temporal logics—particularly linear and signal temporal logic—as expressive specification languages for robot behavior over time. We then discuss principles of formal synthesis, from discrete graph- and game-based approaches to sampling-based motion planning, trajectory optimization, and control-certificate-based synthesis. Finally, we outline challenges in deploying formal synthesis in real-world robotics, emphasizing the interplay between modeling fidelity, computational tractability, and the types of rigorous guarantees that can be achieved.
\end{abstract}


\input{Sections-arxiv/intro}
\input{Sections-arxiv/spec}
\input{Sections-arxiv/synthesis}

\input{Sections-arxiv/extreme}

\input{Sections-arxiv/challenges}
\input{Sections-arxiv/outro}

\bibliographystyle{plainnat}
\bibliography{references}

\end{document}

%% file: Sections-arxiv/intro.tex
\section{Introduction}

Robots are increasingly deployed in the real world—from roads to office environments, public spaces, and even extreme environments that are difficult for humans to reach. In all of these settings, reliable operation is essential, and ensuring that robots perform as expected under a wide range of conditions is crucial. This is where formal methods—broadly defined as rigorous techniques for specification, analysis, and verification—can help.

Traditionally developed within computer science, formal methods include techniques such as temporal logics, model checking, and theorem proving, which enable reasoning about system behavior with provable guarantees. In a broader sense, formal methods also encompass techniques from control theory, such as control barrier functions and reachability analysis, which offer strong guarantees on properties including safety, stability, and robustness. Formal methods are not just a set of tools; they represent a principled way of designing reliable systems.

This manuscript focuses on formal methods specifically in the context of robot planning and control—techniques that generate paths, trajectories, or controllers to move and achieve goals. A first step toward understanding whether a robot behaves as expected is a \emph{specification}: a precise expression of what is required, preferred, and prohibited. In this manuscript, we introduce temporal logics—particularly linear temporal logic and signal temporal logic—as rigorous and expressive specification languages for describing robot behavior over time. Given such a specification, we then turn to \emph{formal synthesis}, the rigorous generation of system behavior that satisfies the specification. Formal synthesis typically provides provable guarantees on whether, and to what extent, the synthesized behavior fulfills the specification. We review several fundamental synthesis principles, including graph- and game-based methods, sampling-based planning, trajectory optimization, and control-certificate-based approaches.

The second part of this manuscript is devoted to the application of formal synthesis in real-world robot systems. While the principles of specification and synthesis can be demonstrated in idealized settings, deploying them in practice introduces several challenges. Real environments are uncertain, dynamic, and often only partially observable, making it difficult to guarantee that specifications are always met. Scalability is another key concern, as the computational cost of synthesis can grow quickly with the size of the system and the complexity of the requirements. Moreover, many assumptions underlying formal synthesis methods—such as perfect models of the robot and its environment—do not hold in practice, requiring strategies to relax or adapt these assumptions while still offering meaningful guarantees. In this manuscript, we overview these challenges and highlight several approaches that balance rigorous guarantees with practical feasibility, illustrating how formal synthesis can be applied effectively in complex robotic scenarios.

\subsection{Objectives}
\begin{itemize}
\item Introduce linear temporal logic and signal temporal logic as expressive specification languages for robot tasks and constraints.
\item Present the fundamental principles of formal synthesis for planning and control.
\item Provide a brief overview of current trends and directions that extend these specification languages and synthesis principles to address challenges in real-world robotics.
\item Discuss the challenges and limitations of applying formal synthesis in practical robotic systems.
\item Illustrate how formal synthesis can be adapted to systems under uncertainty.
\end{itemize}

%% file: Sections-arxiv/spec.tex
\section{Specification}
\label{sec:spec}

Robots operating in the real world face demands that go far beyond reaching a target point or following a collision-free path. Their behavior must respect intricate requirements: visiting locations in a particular order, meeting temporal deadlines, keeping safe distances, reacting to human requests, and adapting to events. Satisfying these requirements begins with their rigorous specification.

A wide range of formalisms have been proposed for this purpose, including planning languages (e.g., PDDL), domain-specific languages, hierarchical structures such as behavior trees, and optimization-based formulations. Each offers different ways of describing what a system should achieve and, in some cases, how it should do so. In this section, we focus on \emph{temporal logics}, a family of specification languages originating in the formal methods research -- research focused on mathematically rigourous techniques for specification, analysis, verification and design of hardware and software systems. Temporal logics, such as linear temporal logic and signal temporal logic provide a natural framework for describing the evolution of system behavior over time, reasoning not just about individiual system states at a particular time instance, but about entire sequences of states.

\begin{example}[Autonomous mobile robot] 
Consider a mobile robot in the environment in Fig.~\ref{fig:running}. The task of the robot may be to periodically survey all rooms to make sure that there are no emergencies and recharge in between the survaillence rounds. It may be asked to bring a cup to the kitchen whenever it finds a dirty one. At the same time, it should not go on the epoxy floor. It may be given explicit temporal guidelines - to visit the kitchen every hour, but all other rooms every other hour. It may be given spatio-temporal guidelines, too, like to stay within the wifi range for as long as possible. Temporal logics can express precisely such requirements with (spatio-) temporal goals, constraints and dependencies.

\begin{figure}
\centering
\includegraphics[width=0.5\linewidth]{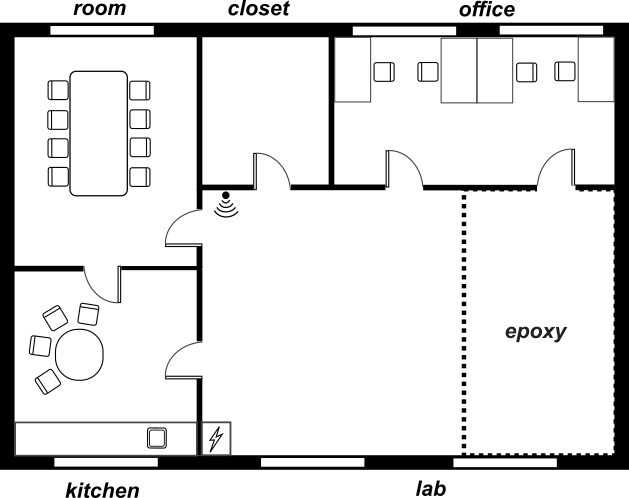}
\caption{A floor plan of an environment with a kitchen, a lab, an office, a closet, and a meeting room (room for short). There is a charging station in one corner of the lab and a wifi router in another.}
\label{fig:running}
\end{figure}

\label{example:running}
\end{example}

\subsection{Linear Temporal Logic}
\label{sec:LTL}

The basic building block of LTL is an \emph{atomic proposition}, which is a statement that holds either true or false. The adjective ``atomic'' indicates that this statement cannot be further broken down into simpler statements. A labeling function $L:X\rightarrow 2^{\AP}$ marks the state of a robot with the subset of atomic propositions that hold true therein, while the rest of the propositions from $\AP$ are false. The labeling function naturally induces partitioning on the robot's state space; the states with the same labels belong to the same partition.

\begin{example}[Autonomous mobile robot cont.] 
 ``The robot is in the lab.'' or ``The robot is within the wifi range.'' are atomic propositions -- properties of relevance and interest in each state of the robot in the environment from Example~\ref{example:running}. $\mathit{AP} = \{\mathit{room, closet, office, kitchen, lab, epoxy, wifi, charger}\}$ is the complete set of atomic propositions. Figure~\ref{fig:running-part} (left) shows the corresponding partitioning of the environment into cells.
 
\begin{figure}
\centering
\includegraphics[width=0.48\linewidth]{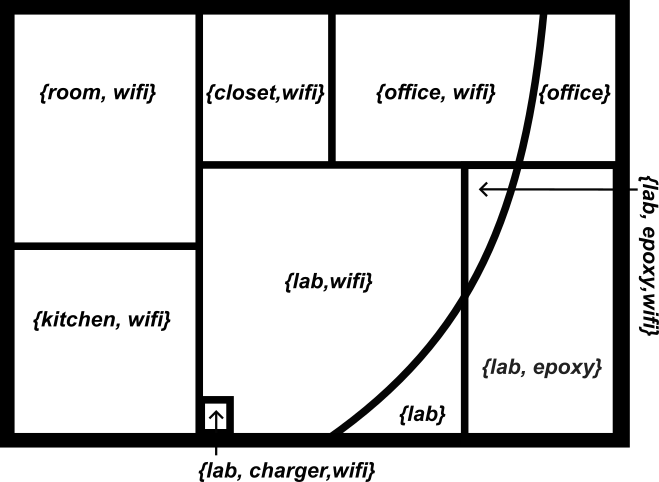}
\includegraphics[width=0.46\linewidth]{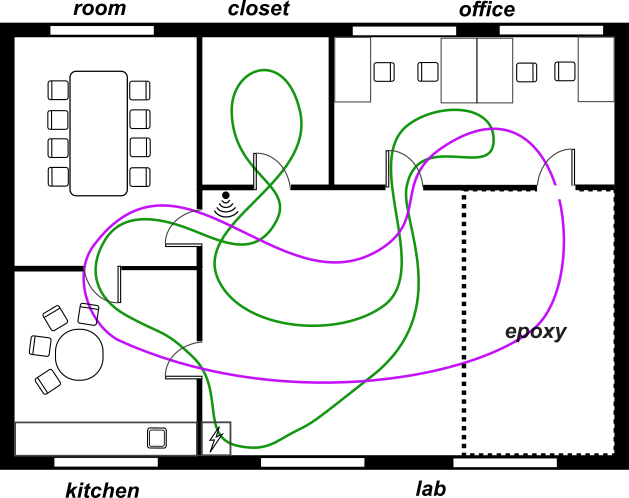}
\caption{(left) Partitioning of the environment based on labeling. A circle centered around the wifi router defines the wifi range, depicted in the figure as the line intersecting the lab and the office. (right) Two periodic trajectories of the robot.}
\label{fig:running-part}
\end{figure}

\label{example:running2}
\end{example}

\begin{definition}[Word]
As the robot moves in the environment, the subset of atomic propositions that hold true changes. Let $t_0t_1t_2\ldots$ denote the times of these changes associated with a robot trajectory $\mathbf{x}(t)$ such that
\begin{itemize}[itemsep=0ex]
    \item $t_0 = 0$, 
    \item $L(\mathbf{x}(t)) = L(\mathbf{x}(t'))$ for all $t, t' \in [t_i, t_{i+1}), i \geq 0$, and
    \item $L(\mathbf{x}(t_i)) \neq L(\mathbf{x}(t_{i+1}))$ for all $i \geq 0$.
\end{itemize}
The infinite discrete sequence $L(x(t_0))L(x(t_1))L(x(t_2))\ldots$ is the \emph{word} produced by the robot trajectory and marks the evolution of properties of relevance and interest. 
\label{def:word}
\end{definition}

\begin{example}[Autonomous mobile robot cont.] The green (the longer) trajectory illustrated in Figure \ref{fig:running-part} (right) starting at the charger is associated with the word $$\{\mathit{lab, charger, wifi}\}\{\mathit{lab,wifi}\}\{\mathit{kitchen,wifi}\}\{\mathit{room,wifi}\}\{\mathit{lab,wifi}\}\ldots.$$
\end{example}

LTL specifies which words are the desired ones. In other words, it distinguishes the desired robot behaviors from the unwanted ones.

\begin{definition}[LTL Syntax]
LTL formulas over a set of atomic propositions $\mathit{AP}$ are defined inductively as follows:
\begin{itemize}[itemsep=0ex]
\item every $a \in AP$ is an LTL formula
\item given that $\varphi_1$ and $\varphi_2$ are LTL formulas, then also $\neg \varphi_1$, $\varphi_1 \wedge \varphi_2$, $\X \varphi_1$, $\varphi_1 \U \varphi_2$ are LTL formulas.
\end{itemize}
\end{definition}

The Boolean operators $\neg$ (not) and $\wedge$ (and) are standard negation and conjunction; with the use of these two, other Boolean operators like disjunction $\vee$ (or), implication $\Rightarrow$ (if), equivalence $\Leftrightarrow$ (if and only if) can also be derived and often occur in LTL formulas. $\X$ (next) and $\U$ (until) are temporal operators that facilitate interpretation of LTL formulas over infinite \emph{words} (also called $\omega$-words), i.e. discrete sequences of symbols from alphabet $\Sigma = 2^\mathit{AP}$, like the ones produced by robot trajectories above. 
We use $\Sigma^\omega$ to denote the set of all infinite words built from symbols of alphabet $\Sigma$ and formally define the satisfaction relation $\models$. 

\begin{definition}[LTL Semantics]
Given an infinite word $\sigma = \sigma_0\sigma_1\sigma_2\ldots \in \Sigma^\omega$, we define that $\sigma$ satisfies $\varphi$ at position $i$ (denoted as $\sigma,i \models \varphi$) as follows:
\begin{align*}
    & \sigma,i \models a, \ a \in \mathit{AP} & \iff & a \in \sigma_i. \\
    & \sigma,i \models \neg \varphi & \iff  &  \sigma,i \not\models \varphi. \\
    & \sigma,i \models \varphi_1 \wedge \varphi_2 &  \iff  & \sigma,i \models \varphi_1 \text{ and } \sigma,i \models \phi_2. \\
    & \sigma,i \models \X  \varphi &  \iff & \sigma,{i+1} \models \varphi. \\
    & \sigma,i \models \varphi_1  \U  \varphi_2 &  \iff  & \exists j, i \leq j: \sigma, j \models \varphi_2 
    \text{ and } \forall k, i \leq k < j: \sigma, k \models \varphi_1. 
\end{align*}
\end{definition}

Intuitively, $\X \varphi$ represents that $\varphi$ should hold in the \emph{next} step and $\varphi_1 \U \varphi_2$ that $\varphi_1$ holds \emph{until} $\varphi_2$ holds true. 
Two of the most useful LTL operators are derived operators $\F$ (in future, eventually, also sometimes denoted by $\diamond$) and $\G$ (globally, always, also sometimes denoted by $\square$) defined as follows:
\begin{align*}
    \F \varphi & \equiv \mathit{true}\U \varphi \\
    \G \varphi & \equiv \neg \F \neg\varphi
\end{align*}
$\F \varphi$ says that $\varphi$ needs to hold at some time in the future while $\G \varphi$ that $\varphi$ holds globally along the infinite word. 
In other words, $\F$ expresses reachability and $\G$ safety.

\bigskip

By combining and nesting the operators, we get LTL formula patterns representing a variety of interesting specification types, such as:
\begin{itemize}
\item $\G \F \varphi$: always, there is a future point when $\varphi$ holds; or $\varphi$ is periodically revisited. This type of formulas are suitable for expressing, e.g., periodic surveillance properties.
\item $\F \G \varphi$: from some future point on, $\varphi$ holds forever. This type of formulas can be used to express properties related to stability or convergence.
\item $(\neg \varphi_3 \U \varphi_2) \wedge (\neg \varphi_2 \U \varphi_1)$: $\varphi_3$ cannot hold until $\varphi_2$ holds, which cannot hold until $\varphi_1$ holds. In other words, this formula expresses a sequencing requirement: first $\varphi_1$ needs to happen, after that $\varphi_2$ and only after that $\varphi_3$. 
\item $\G(\varphi_1 \Rightarrow \F \varphi_2)$: whenever $\varphi_1$ (a request) happens, $\varphi_2$ (a response) needs to come afterwords. 
\end{itemize}

\begin{example}[Autonomous mobile robot cont.] 
$$\G\F\mathit{kitchen} \, \wedge \, \G (\mathit\mathit{kitchen} \Rightarrow \X \mathit{room}) \, \wedge \, \G\F\mathit{recharge} \, \wedge \, \G \mathit{wifi} \, \wedge \, \neg \G \mathit{epoxy}$$ tells the robot to periodically survey the kitchen, to visit the meeting room right after every visit to the kitchen, to periodically recharge, to always stay within the wifi range and avoid the epoxy floor. The green (the longer) trajectory in Figure \ref{fig:running-part} (right) satisfies this specification, while the purple (the shorter) trajectory does not as the robot goes outside the wifi range for a little while.
\end{example}

\begin{remark} In Definition \ref{def:word}, we have presented one of the ways a word can be constructed; however, there are other common ways used in the literature. The robot's environment can be partitioned, e.g., via triangulation into finer cells that respect the labeling. Reading the labels of the cells as the robot travels through them gives another word. Yet another approach is to consider discrete time-sampled trajectories and read the labels every $\Delta t$.
\end{remark}

So far, we have introduced LTL as a specification language for properties related a robot's location in its environment, where changes to the truth values of atomic propositions are triggered solely by the robot's movement. However, atomic propositions can also refer to the state of the robot's (dynamic) environment. For instance, atomic proposition ``There is a cup on the table'' evaluates the same regardless of the robot's location, but changes depending on the changes in the environment, such as whether a person moves a cup away from the table. Another example is an atomic proposition ``The robot is at a safe distance from a human''. The value and the change of the value of this atomic proposition depends on both the robot and the humans. Broadening the view of atomic propositions allows to broaden the use LTL to express tasks beyond the above examples. For example, LTL was recently used as a specification language for object rearrangement~\citep{vasilopoulos-manipulation2021}, soup-scooping~\citep{wang2022cooking}, or wiping~\citep{nawaz-interactive2024}.

\subsection{Signal Temporal Logic}
\label{sec:STL}

Signal Temporal Logic (STL)~\citep{maler-monitoring} expands LTL in two distinct ways: First, it is interpreted over continuous signals rather than discrete sequences and second, it allows explicit quantitative time constraints. 

Suppose that $\mathbf{x}:\mathbb{R}_{\ge 0} \rightarrow X \subseteq \mathbb{R}^n$ is an $n$-dimensional continuous-time signal, such as the robot trajectory in Example \ref{example:running}. The basic building block of STL is a real-valued \emph{predicate function} $\mu:X \rightarrow \mathbb{R}$, for example the robot's distance to the nearest obstacle, and a \emph{threshold predicate} $p_\mu = \mu(x) \geq 0$, for example what the desired distance to the nearest obstacles is. Similarly to LTL, STL is then built using Boolean and temporal operators. 

\begin{definition}[STL Syntax]
\label{def:stl}
    STL formulas are defined inductively as follows:
    \begin{itemize}
\item a threshold predicate $p_\mu$ is an STL formula
\item given that $\varphi_1$ and $\varphi_2$ are STL formulas, then also $\neg \varphi_1$, $\varphi_1 \wedge \varphi_2$, $\varphi_1 \UI \varphi_2$ are STL formulas.
\end{itemize}
\end{definition}

 The until operator $\mathcal{U}_I$ is explicitly constrained by an open, closed, or half-open subinterval $I$ of the interval $[0,\infty)$ with endpoints in $\mathbb R \cup \{\infty\}$. Formula $\varphi_1 \UI \varphi_2$ specifies that $\varphi_1$ should hold until, within $I$, $\varphi_2$ holds. Similarly as for LTL, additional useful temporal operators can be defined as $\FI \varphi = \top \, \UI \, \varphi$ ($\varphi$ holds eventually within $I$), and $\GI \varphi = \neg \, \FI \, \neg \varphi$ ($\varphi$ holds at all times within $I$). If $I = [0,\infty)$, the subscripts can be omitted, leading to the original operators $\U$, $\F$ and $\G$.

\begin{definition}[STL Semantics] We define that $\mathbf x$ satisfies 
 $\varphi$ at time $t$ ($\mathbf x,t \models \varphi$) as follows
 \begin{align*}
     \mathbf{x},t & \models p_\mu & \iff & \mu(\mathbf{x}(t)) \geq 0. \\
     \mathbf{x},t  & \models \neg \varphi  & \iff & \mathbf{x},t \not\models \varphi. \\
     \mathbf{x},t  & \models \varphi_1 \wedge \varphi_2  & \iff & \mathbf{x},t  \models \varphi_1 \text{ and } \mathbf{x},t  \models \phi_2. \\
     \mathbf{x},t  & \models \varphi_1  \UI  \varphi_2  & \iff & \exists t'' \in t+I: \mathbf{x},t'' \models \varphi_2 
    \text{ and } \forall t' \in [t,t''): \mathbf{x},t'  \models \varphi_1. 
\end{align*}
\end{definition}

STL can also be evaluated quantitatively using \emph{robustness}, which intuitively captures the degree of satisfaction or violation of an STL formula \citep{donze2010robust}. \emph{Spatial robustness} $\rho$ is based on the value of the predicate function, i.e., it reflects the robustness with respect to satisfying the threshold predicates:

\begin{definition}[STL Space Robustness]
\begin{align*}
    \rho_{p_\mu}(\mathbf x, t) &= \mu(\mathbf x(t)), \\
    \rho_{\neg \varphi}(\mathbf x, t) &= -\rho_{\varphi}(\mathbf x, t), \\
    \rho_{\varphi_1 \land \varphi_2}(\mathbf x, t) &= \min(\rho_{\varphi_1}(\mathbf x, t),\rho_{\varphi_2}(\mathbf x, t)), \\
    \rho_{\varphi_1 \mathcal{U}_I \varphi_2}(\mathbf x, t) &= \max_{t'' \in t+I}\big(\min(\rho_{\varphi_2}(\mathbf x, t''),\min_{t'\in[t,t'']}\rho_{\varphi_1}(\mathbf x, t') ) \big) 
\end{align*}
In particular for the $\F$ and $\G$ operators, we have
\begin{align*}
\rho_{\FI \varphi} (\mathbf x,t) &= \max_{t'\in t+I} \rho_\varphi(\mathbf x,t')
     \\
    \rho_{\GI \varphi}(\mathbf x,t) &= \min_{t'\in t+I} \rho_\varphi(\mathbf x,t')
\label{eq:stl_space_robustness_FG}
\end{align*}
\label{eq:stl_space_robustness}
\end{definition}

\emph{Temporal robustness}, on the other hand, represents robustness against delays and advancements, see e.g. \citep{donze2010robust,rodionova-temporalSTL2023} for definitions.

\begin{example}[Autonomous mobile robot cont.]
    Consider a threshold predicate $p_\mu = \mathit{dist(robot,wifi)} \leq \mathit{rad}$, where $\mathit{dist(robot,wifi)}$ is the distance between the robot and the wifi router and $\mathit{rad}$ is the wifi radius. STL formula $\G p_\mu$ is satisfied for the green trajectory in Figure \ref{fig:running-part} (right) and the spatial robustness is positive. On the other and, it is violated for the purple trajectory and the spatial robustness is negative, measured at the furthest point of the trajectory from the wifi router.
\end{example}

\subsection{Other Temporal Logics}
\label{sec:otherlogics}
\paragraph{Metric temporal logic} LTL and STL differ not only in their expressiveness, but also in the algorithmic machinery supporting its synthesis as we will further discuss in Section~\ref{sec:principles}. LTL is frequently coupled with analysis of discrete graphs and games, and verification and synthesis problems are generally decidable. Due to its continuous semantics, STL does not enjoy benefits of the same approaches and verification and synthesis problems are generally undecidable.
Certain temporal logics offer a compromise. \emph{Metric Temporal Logic} (MTL) can be viewed as an extension of LTL with explicit time intervals and interpretation over timed words; or it can be viewed as a restricted fragment of STL where threshold predicates are replaced with atomic propositions. Similar as LTL, MTL allows for the use of discrete graphs and games. 

\paragraph{Expressive temporal operators} Various robotics scenarios inspired the creation of new logics and new temporal operators to offer intuitive ways to express interesting tasks combined with suitable algorithmic support for synthesis. For example, time window temporal logic introduces \emph{concatenate}, \emph{within} and \emph{hold} operators to express, e.g., that a robot should service certain locations in a sequence, each for a certain duration within a certain interval \citep{cristi-twt2017}. Weighted STL in turn allows to specify importance of priorities of subformulae \citep{noushin-wstl2021}. 

\paragraph{Logics to deal with uncertainty}  
Probabilistic extensions of temporal logics like PrSTL have been introduced to deal with the stochastic nature of many real-world robotic systems that are subject to omnipresent uncertainty \citep{sadigh2016safe}. For example, \cite{safaoui2020control} proposed risk-based STL built from risk constraints on atomic predicates interpreted in stochastic states of a stochastic process.

\paragraph{Branching-time logics} So far, we have discussed linear logics, whose formulas are interpreted over individual sequences. Branching-time logics like Computational Tree Logic (CTL) are interpreted over trees of possible executions. This viewpoint has been useful especially in context of stochastic systems, where Probabilistic CTL (PCTL) can express desired properties of robots with imprecise sensors or actuators \citep{morteza-pctl2012}.

\subsection{Designing Temporal Logic Specifications}

There are two main ways to design specification in temporal logic: The first one involves explicit input -- either directly in the form of the temporal logic formula, or a more user-friendly means that maps onto a temporal logic formula.  For instance, LTLMoP supports formulating robot goals in structured English \citep{finucane2010ltlmop}. Other work considers a graphical user interface for LTL motion and mission planning \citep{ltlviz}. 
The topic of translating input in natural language into LTL or STL has recently gained popularity due to advances in LLMs 
including works that consider guarantees on the correctness of the translation~\citep{wang2025conformalnl2ltltranslatingnaturallanguage}. 

The second way to design temporal logic specifications is learning from examples of ``good'' (and possibly also ``bad'') behaviors. 
For example, \cite{ltllfd-shah2018}  and \cite{chou2022learning} learn LTL formulas from demonstrations. Other works have focused on learning STL formulas or parameters of their threshold predicates
\citep{bombara21,alexis-multiclass2022,aasi-2023}.

\subsection{Main takeaways}

\begin{itemize}
\item Rigorous, unambiguous specification is essential: it enables imposing requirements and constraints, and precisely evaluating whether—and to what extent—they are satisfied.
\item Different logics suit different needs: LTL is well-suited for high-level, discrete tasks and constraints, while STL naturally supports continuous dynamics and quantitative requirements. Other temporal logics extend these ideas with additional operators, or semantics supporting probabilistic reasoning.
\item Specifications can be created in multiple ways: explicitly through logical formulas or user-friendly interfaces (including natural language), or implicitly by learning from demonstrations and data.
\end{itemize}

%% file: Sections-arxiv/synthesis.tex
\newpage

\section{Principles of model-based formal synthesis}

\label{sec:principles}

\emph{Formal synthesis} refers to the automated generation of a system or a system behavior from a formal specification, such as a temporal logic formula. Typical for formal synthesis are provable guarantees on whether or how well the synthesized system or system behavior satisfies the specification.

In this section, we focus on the offline model-based formal synthesis of a single agent \emph{behavior}, which means a robot task plan, a motion plan, a policy, a trajectory, or a control strategy, depending on the model considered. 
We distinguish three types of formal synthesis problem: a \emph{feasibility} problem and two \emph{optimization} problems -- one where the satisfaction of a formal specification is considered as a constraint and one where the degree of specification satisfaction is a part of the optimization objective. 

\begin{problem}[Feasibility Problem]
\label{prob:feasibility}
Given a robot model and a temporal logic specification $\varphi$, find a behavior that guarantees the satisfaction of $\varphi$.
\end{problem}

The feasibility problem is also known as the correct-by-design synthesis problem; one may also be interested in finding a correct-by-design behavior that is optimal, e.g., in terms of time or energy, giving rise to the first type of optimization problem:

\begin{problem}[Constrained Optimization Problem]
\label{prob:constrained_optimization}
Given a robot model and a temporal logic specification $\varphi$, find a behavior that guarantees the satisfaction of $\varphi$ and optimizes an objective function. 
\end{problem}

In the third type of formal synthesis problem, satisfaction of the temporal logic specification is not enforced as a hard constraint but instead forms part of the objective function. This formulation is particularly useful when the specification cannot be fully satisfied and we seek to minimize specification violation or to maximize robustness.

\begin{problem}[Optimal Satisfaction Problem]
\label{prob:optimal_satisfaction}
Given a robot model and a temporal logic specification $\varphi$, find a behavior that maximizes the degree of satisfaction of~$\varphi$.
\end{problem}
 
The specifics of the problem formulation depend on the model class -- ranging from finite discrete-time transition systems to continuous-time dynamical systems -- and the nature of the logic. In the following, we will introduce basic principles of formal synthesis suited for different model classes, roughly following the traditional planning hierarchy: from abstract, high-level discrete models suited for long-horizon planning, through motion planning, to low-level continuous models such as differential equations used for control.

\subsection{Graph- and game-based synthesis}
\label{sec:graph}
Graph and game-based synthesis approaches generally rely on modeling the system through some sort of discrete state-transition model. In the most basic cases, the model is a finite deterministic labeled transition system, the specification is an LTL formula, and the aim is to solve the feasibility problem. 

\subsubsection{Synthesis problem for deterministic transition systems}

\begin{definition}[Deterministic Transition System]
A labeled deterministic transition system (DTS) is a tuple $\T=(S, \mathit{Act}, R, \sinit, \mathit{AP}, L)$, where 
    $S$ is a finite set of states, 
    $\mathit{Act}$ is a finite set of actions, 
    $R : S \times \mathit{Act} \xrightarrow{} S$ is the transition function,
    $\sinit \in S$ is the initial state,
    $\mathit{AP}$ is a finite set of atomic propositions and
    $L : S \xrightarrow{} 2^{\mathit{AP}}$ is a labeling function. 
\end{definition}

A \emph{control strategy} in DTS $\T$ is an infinite sequence of actions $\pi = a_0a_1a_2\ldots$ from $Act$. Following such control strategy results in a unique path $s_0\xrightarrow{a_0}s_1\xrightarrow{a_1}s_2\xrightarrow{a_2}\ldots$, where $s_1=\sinit$ and $s_{i+1}=R(s_i,a_i)$, for all $i\geq 0 $. The path is associated with a unique word $\sigma_\pi = L(s_0)L(s_1)L(s_2)\ldots$. 

One way to create a DTS is from the partitioning induced by labeling as discussed in Sec. \ref{sec:LTL}: each partition represents a state of the DTS and each transition represents the robot's capability to navigate between two partitions. 

\begin{figure}
\centering
        \begin{tikzpicture}[shorten >=1pt,node distance=3cm,on grid,auto] 
            \node[state] (top1)   {$s_1$};
            \node[] (top1label) [above = 0.8cm of top1] {$\{ \mathit{room,wifi} \}$};
            \node[state] (top2) [ right =of top1] {$s_2$};
            \node[] (top2label) [above = 0.8cm of top2] {$\{ \mathit{closet,wifi} \}$};
            \node[] (top2fake) [right=1.5cm of top2]{};
            \node[state] (top3) [ right =of top2] {$s_3$};
            \node[] (top3label) [above = 0.8cm of top3] {$\{ \mathit{office,wifi} \}$};
            \node[state] (top4) [ right =of top3] {$s_4$};
            \node[] (top4label) [above = 0.8cm of top4] {$\{ \mathit{office}\}$};
            \node[state] (mid2) [ below  = 2cm of top2fake] {$s_5$};
            \node[state] (mid3) [ below right = 2.2cm of top3] {$s_6$};
            \node[] (mid2label) [left = 1.7cm of mid2] {$\{ \mathit{lab,wifi}\}$};
            \node[state] (bot1) [ below = 3.3cm of top1] {$s_7$};
            \node[] (mid3label) [above = 0.7cm of mid3] {$\{ \mathit{lab,epoxy,wifi} \}$};
            \node[state,initial left, initial text=] (bot2) [ below  = 4cm of top2] {$s_8$};
            \node[] (bot2label) [below = 0.8cm of bot2] {$\{ \mathit{lab, charger,wifi}\}$};
            \node[] (bot1label) [below = 0.8cm of bot1] {$\{ \mathit{kitchen,wifi} \}$};
            \node[state] (bot4) [ below = 4cm of top4] {$s_{10}$};
            \node[state] (bot3) [left = 2.5cm of bot4] {$s_9$};
            \node[] (bot3label) [below = 0.8cm of bot3] {$\{\mathit{lab}\}$};
            \node[] (bot4label) [below = 0.8cm of bot4] {$\{\mathit{lab,epoxy}\}$};
            \path[->]
            (top1) edge node [pos=0.2, left] {} (bot1)
            (top2) edge node [pos=0.2] {} (mid2)
            (top3) edge node [pos=0.2] {} (mid2)
            (top4) edge node [pos=0.2] {} (bot4)
            (top3) edge node [pos=0.2]{} (top4)
            (top4) edge node [pos=0.2, above]{} (top3)
            (top1) edge node [above,pos=0.2] {} (mid2)
            (bot1) edge node [pos=0.2]{} (top1)
            (bot2) edge node [pos=0.2,right]{} (mid2)
            (bot3) edge node [pos=0.2,right]{} (mid2)
            (bot4) edge node [pos=0.2]{} (mid3)
            (bot4) edge node [pos=0.2, right]{} (top4)
            (bot1) edge node [pos=0.2]{} (mid2)
            (bot4) edge node [pos=0.2]{} (bot3)
            (bot3) edge node [pos=0.2,below]{} (bot4)
            (mid3) edge node [pos=0.2] {} (mid2)
            (mid3) edge node [pos=0.2, below]{} (bot4)
            (mid2) edge node [pos=0.2]{} (top2)
            (mid2) edge node [pos=0.2]{} (top3)
            (mid2) edge node [pos=0.2]{} (top1)
            (mid2) edge node [pos=0.2]{} (bot1)
            (mid2) edge node [pos=0.2]{} (bot2)
            (mid2) edge node [pos=0.2]{} (bot3)
            (mid2) edge node [pos=0.2]{} (mid3)          
            ;       
        \end{tikzpicture}
        \caption{A DTS. The states are named $s_1$-$s_{10}$ and labeled with a subset of atomic propositions. The initial state $s_8$ is marked with an incoming arrow. The edges represent transitions between states; for simplicity, we have omitted the action names in the figure. }
        \label{fig:DTS}
\end{figure}
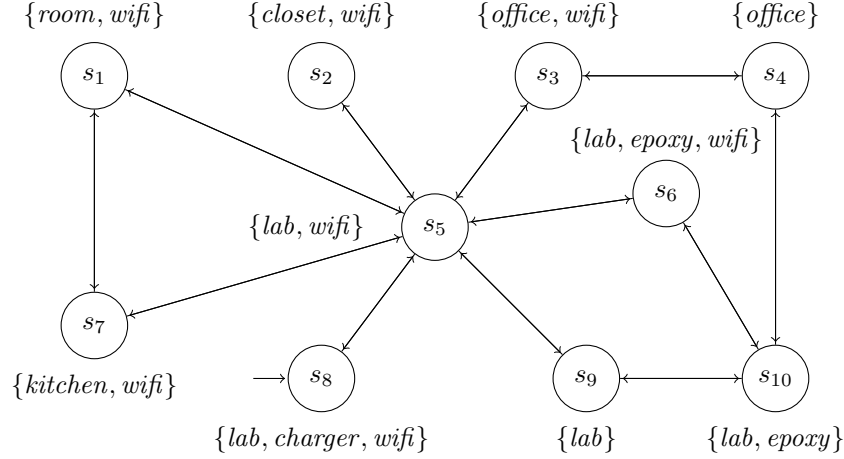

\begin{example}[Autonomous mobile robot cont.]
The DTS built from the partitioning in Fig. \ref{fig:running-part} (left) for a fully actuated robot is illustrated in Fig. \ref{fig:DTS}.
\end{example}

Building a DTS from the aforementioned partitioning is not the only way to build a DTS model of the robot in its environment. \cite{plaku-2012-LTL} discusses how to bridge the gap between this type of high--level discrete planning and motion planning. In turn, \cite{kress2018synthesis} present an overview of techniques to create abstractions (DTS and beyond) of the physical system. 

{Given a DTS and an LTL formula $\varphi$ over the atomic propositions of the DTS, the feasibility problem is to find a control strategy $\pi$ such that the corresponding word $\sigma_\pi$ satisfies $\varphi$.} 

\subsubsection{Automata-based approach}

To solve the problem, a so-called automata-based approach heavily inspired by LTL model checking procedure \citep{baier2008principles} takes four steps: (1) the LTL formula is translated to a B\"uchi automaton, (2) the DTS is combined with the B\"uchi automaton into a product automaton, (3) the product automaton is treated as a graph and analyzed to find a certain type of path in it, and (4) the path in the product automaton maps onto a control strategy in the DTS. The details follow.

\begin{definition}[B\"uchi Automaton] 
A B\"uchi Automaton (BA) over the alphabet $\Sigma$ is a tuple $\A=(Q, \Sigma, \delta, \qinit, F)$, where 
 $Q$ is a finite set of states, 
 $\Sigma$ is a finite set of symbols,
 $\delta : Q \times \Sigma \xrightarrow{} 2^Q$ is the transition function,
 $\qinit$ is the initial state, and
 $F \subseteq Q$ is a  set of \textit{accepting} states.
\end{definition}

A BA takes as an input an infinite word from $\Sigma^\omega$. A run of $\A$ over $\sigma = \sigma_0\sigma_1\sigma_2\dots$ is a sequence of states $r_\sigma = q_0q_1q_2\dots$ such that $q_0 = \qinit$, $q_{i+1} \in \delta(q_i, \sigma_i)$, for all $i \geq 0$. If there exists an \emph{accepting run} over $\sigma$, i.e. a run $r_\sigma$ that intersects $F$ infinitely many times, then $\sigma$ is \emph{accepted} by $\A$.

A key insight is that any LTL formula $\varphi$ can be algorithmically translated into a corresponding B\"uchi automaton $\A_\varphi$, meaning that word $\sigma \in \Sigma^\omega$ satisfies $\varphi$ if and only if it is accepted by $\A_\varphi$. The size of $\A_\varphi$ is generally exponential with respect to the length of the LTL formula, but in practice it is usually much smaller.

At this point, the feasibility problem is: Given a DTS and a BA $\A_\varphi$, find a control strategy $\pi$ such that the corresponding word $\sigma_\pi$ is accepted by the BA. To capture the intersection of all words of the DTS and all words accepted by the BA, we design a product automaton:

\begin{definition}[Product Automaton]
Given a DTS $\T=(S, \mathit{Act}, R, \sinit, \mathit{AP}, L)$, and a BA $\A_\varphi=(Q, \Sigma = 2^\AP, \delta, \qinit, F)$, the product automaton is a BA $\Prod = \T \otimes A = (S\times Q, Act, \delta_\Prod, (\sinit,\qinit), S\times F) $, where 
$(s',q') \in \delta_\Prod((s,q),a)$ if and only if $R(s,a)=s'$ and $q' \in \delta(q,L(s))$.
\end{definition}

 In words, an action taken in the transition system leads both to a new state (a transition in the TS) and an observation of the APs (a transition in the BA). The product automaton ensures that the transitions in the BA are restricted by the 'physical' limitations of the TS.
This construction ensures two key properties: an accepting run in the product automaton maps onto a control strategy in $\T$ resulting in satisfaction of $\varphi$ and vice versa, if there is a control strategy in $\T$ leading to satisfaction of $\varphi$, there is a corresponding accepting run in the product automaton. Furthermore, a useful observation is that if there exists an accepting run in the product automaton, there also exists one in a prefix-suffix form $r_\mathit{pref}(r_\mathit{suf})^\omega$, where a finite prefix is followed by an infinite repetition of a finite suffix. 

When we view the product automaton as a graph with $S$ being the set of vertices and $R$ defining the edges between the vertices, the remaining steps boil down to finding a so-called lasso -- a finite path followed by a cycle containing an accepting state via graph search algorithms. By projection of the lasso onto states and actions of the DTS, we immediately obtain the desired control strategy in the prefix-suffix form. If no lasso is found in the product automaton, then there is no control strategy guaranteeing the satisfaction of the LTL specification. 

\subsubsection{Extensions and variations}

\paragraph{Uncertainty} The automata-based synthesis approach can be extended to non-deterministic transitions systems (NDTSs) and Markov Decision Processes (MDPs) to address uncertainty coming from e.g., the uncontrollable parts of the environment, disturbances, sensors or actuators imprecision, or modeling errors. The extension comes with certain technical challenges: in general, the BA used in the product cannot be non-deterministic. While some LTL formulas translate to a deterministic BA, for some others, an alternative like deterministic Rabin Automaton (DRA) has to be used with double-exponential size with respect to the length of the LTL formula. The analysis of the product automaton then differs. The product automaton is now viewed as a game on a graph (with the non-determinism being viewed as the adversarial player and stochasticity as a 1/2 player). Correspondingly, one needs to find a winning strategy \citep{kloetzer2008dealing, ltlmdp}. The resulting control strategy for the NDTS or MDP is limited history-dependent and often comes in the form of, e.g., an input-output automaton that returns the next action to be taken depending on the current state. Automata-based synthesis for Partially Observable MDPs (POMDPs) with LTL specifications is especially computationally challenging \citep{
chatterjee2015qualitative}. 

\paragraph{Scalability} 
A successful strategy to deal with the high computational demands associated with synthesis for full LTL stemming from the exponential translation of a specification to a corresponding automaton, is to focus on fragments of LTL. For example, control strategies satisfying syntactically co-safe LTL (sc-LTL) formulas take the form of a prefix followed by an arbitrary suffix. This observation turns the infinite-horizon problem described above into a finite-horizon problem, since the satisfaction of any sc-LTL formula can be decided by a deterministic finite automaton \citep{bhatia-sampling2010}.
General Reactivity (1) (GR(1)) is another fragment of LTL useful to express robot tasks \citep{hadas-tro09}. In GR(1) synthesis, the system is not explicitly modeled. Rather than that, the assumptions on the robot and its environment are captured through an LTL formula $\varphi_e$ and the aim is to satisfy $\varphi_e \implies \varphi_g$, where $\varphi_g$ represents the goal. While GR(1) is more restrictive than full LTL, it allows to avoid the construction of the B\"uchi automaton and hence comes with a more favorable computational complexity. 

Further strategies to improve scalability to larger problems include those that avoid analyzing the entire product automaton at once, e.g., by planning over a receding horizon \citep{nok-rh2012}, or by sampling-based search \citep{Kantaros-samplingmultiTRO2019}. Often, the price paid for efficiency is weaker (probabilistic) completeness guarantees.

\paragraph{Strategy non-existence} Synthesis for the feasibility problem may not always find a successful strategy for several reasons: given specification might not be realizable, the modeling or the imposed assumptions might be too conservative, or the chosen synthesis approach might not be complete. In such a case, one could look for a close-to-correct-by-design controller instead. For example, \citet{fainekos-2011revising} proposed to revise the specification so that it becomes feasible. 
Another line of work suggested quantitative semantics for LTL to measure the level of violation and turned Problem~\ref{prob:feasibility} into Problem  \ref{prob:optimal_satisfaction} \citep{tumova2013least, maly-2013}.
Other work considered adding assumptions to enable guarantees \citep{li-2011-mining} or modifying the system, such as adding on robot's skills \citep{pacheck-2023-repair}.

\paragraph{Optimality} With some effort, automata-based synthesis can be adapted to address Problem \ref{prob:constrained_optimization}. Since LTL is interpreted over infinite words, one needs to associate meaningful cost of an infinite control strategy. For example, \cite{steve-ijrr2011} proposed to extend the transition system with weights, and minimize the worst-case cumulative weight between two subsequent satisfactions of an atomic proposition, marking e.g., a robot's base station. The weights are carried over to the product automaton, where one searches for an optimal lasso. A similar method was in fact used to cope with Problem \ref{prob:optimal_satisfaction} with least-violation LTL semantics \citep{tumova2013least}. 

\paragraph{Specification expressiveness} MITL specifications can be translated into timed automata, enabling the extension of automata-based synthesis techniques to explicitly incorporate timing constraints. Naturally, the underlying system model must also include temporal information, which can take the form of timed automata or, more generally, hybrid or dynamical systems that are abstracted into finite models \citep{bouyer-2006mtl, liu-2014mtl}. STL is interpreted over continuous signals and is therefore particularly well-suited for specifying properties of dynamical systems. Consequently, automata-based approach -- if used at all -- is deployed to complement or guide control synthesis for dynamical systems \citep{lindemann-2020acc, gundana-2021stl,ho-2022cdc}.

\paragraph{Dynamical models} 
The techniques described above are characterized by building a discrete abstraction of the underlying system. Another line of game-based approaches uses directly dynamical or hybrid models in combination with dynamic programming techniques. For instance, \cite{chen2018signal} draw a connection between temporal logic operators and reachability operations. Robustness guarantees upon execution come at the cost of (approximately) solving a high-dimensional Partial Differential Equation (PDE).

\subsection{Synthesis via sampling-based motion planning}

Sampling-based motion planning algorithms, such as RRT and PRM, have proven highly effective for traditional motion planning problems, particularly in high-dimensional continuous spaces where explicit representations of the configuration space are computationally prohibitive. However, they are not directly applicable to specifications expressed in temporal logic, which require reasoning about sequences of events and temporal relations in addition to geometric feasibility.
In this section, we outline how principles of sampling-based motion planning can be extended to handle temporal logic specifications, starting from the feasibility problem:
Given a configuration space $\mathcal X \subseteq \mathbb R^n$, an initial configuration of the robot $x_\mathit{init} \in \mathcal X$, a partitioning of the configuration space and a labeling function that assigns a subset of atomic propositions $\AP$ to each partition, find a path originating at $x_\mathit{init}$ that satisfies a given LTL formula $\varphi$ over $\AP$.

\subsubsection{Multi-layered planning}

One approach addresses the problem via a multi-layered approach \citep{bhatia-sampling2010}. A discrete abstraction and planning layer follows the principles described in Sec.~\ref{sec:graph} and interacts with low-level sampling-based planning. Together, the two layers ensure that the resulting paths satisfy both the temporal logic specification and the system’s feasibility constraints.

\subsubsection{Adapting rapidly exploring random trees and graphs}

Another approach directly adapts incremental sampling-based algorithms, particularly those based on rapidly exploring random trees or graphs \citep{cristi-sampling2013, luo-2021-abstractionfree}. The core idea is to view the incrementally updated tree or graph as a discrete state-transition model and apply the principles from Sec.~\ref{sec:graph}. Namely, the LTL specification is translated into a BA, a product automaton between the incrementally updated tree or graph and the BA is maintained and iteratively analyzed to recognize whether the LTL specification is satisfied. Note that LTL is traditionally interpreted over infinite time horizon while motion planning often focuses on finding finite paths. As a remedy, either the specification needs to be interpretable over finite horizons, such as co-safe fragment of LTL, or the motion planning algorithms has to admit cycles, such as Rapidly Exploring Random Graphs (RRG).
Generally, this approach does not require to build full abstraction of the system and preserves desirable properties of the underlying motion planning algorithm, such as probabilistic completeness and/or asymptotic optimality.

\subsubsection{Extensions and variations}

\paragraph{Optimality} A popular way to address Problem \ref{prob:constrained_optimization} specifications is to adapt the asymptotically optimal RRT* motion planning algorithm in a similar way as RRT and RRG were adapted for the feasibility problem \citep{kyunghoon-cost2017, ZHANG2020105591}. Problem \ref{prob:optimal_satisfaction} -- the problem of maximizing temporal logic satisfaction has been considered, too. The main idea of many solutions is to use quantitative semantics, such as robustness for STL, as the optimization function that guides the ``rewiring'' procedure in RRT* \citep{luis-leastviolating203, cristi-samplingmaxsat2017, alexis-realtimerrt2023}.
Both mentioned types of RRT* adaptation often inherit asymptotic optimality properties.

\paragraph{Scalability}
To improve scalability and computational tractability, many methods adopt strategies similar to those used in standard sampling-based motion planning—such as hierarchical planning, guided or biased sampling, and pruning—while additionally incorporating reasoning about the specification.

A major contributor to high computational demands arises from kinodynamic constraints. To incorporate these, many works expand on the multi-layered planning approach, e.g. by incorporating accurate simulations of dynamics, guiding search for kinodynamic paths by geometric paths, or use a combination of trajectory optimization and closed-loop controllers to derive dynamically feasible trajectories  \citep{plaku-sampling2014, chatrola2025kinodynamic, marchesini2025sampling}. Another helpful strategy focuses on biasing sampling and guiding the steering procedure, e.g., towards increased satisfaction of the task \cite{cristi-samplingmaxsat2017}.

\paragraph{Uncertainty} Another challenging factor is uncertainty. Recent works often adopt the multi-layered approach, where uncertainty is handled separately at the high-level discrete planning layer and/or at the low-level planning \citep{9293004, ho2023planning}. 

\subsection{Synthesis via trajectory optimization}
\label{sec:trajopt}

Trajectory optimization is a core methodology in robotics, offering a principled way to generate continuous, dynamically feasible motions that satisfy task and physical constraints. 
When tasks involve complex spatio-temporal requirements, combination with temporal logics, and particularly STL becomes useful. In this section, we focus primarily on Problem \ref{prob:optimal_satisfaction}, where, given an Ordinary Differential Equation (ODE) model of the robot and its environment, and an STL specification, we look for a control signal $u(t)$ which ensures maximal STL robustness. 

The main challenge is encoding the specification into constraints and optimization function. While the underlying system dynamics and constraints can often be expressed in smooth, continuous forms, this is not the case for spatio-temporal tasks. Specifically, STL robustness
in Def.~\ref{eq:stl_space_robustness} uses $\min$ and $\max$ operations, which, even when defined over convex and differentiable predicates, are non-convex and non-smooth.
As a result, optimization problems involving these operators are inherently non-convex and non-smooth.
\subsubsection{Mixed-integer programming formulation}
    The first approach to tackle the above challenge is to formulate a Mixed-Integer Programming (MIP) problem, which allows for both continuous and integer variables (often binary). The latter ones can be used to embed the STL temporal operators via e.g., the big-M method ~\citep{raman2014model}. 
    
    For example, consider the STL specification $\varphi = \F_{I}(\mu(\mathbf x(t))\geq0)$.
    The associated spatial robustness is $\rho_{\varphi} = \max_{t\in I}(\mu(\mathbf x(t)))$ and Problem \ref{prob:optimal_satisfaction} translates to maximizing $\rho_\varphi$ under the following set of mixed-integer constraints
    \begin{align*}
        &\rho_{\varphi} \geq \mu(\mathbf x(t)), \quad \forall t \in I, \\
        &\rho_{\varphi} \leq \mu(\mathbf x(t)) + M(1-\delta_t), \quad \forall t \in I, \\
        &\sum_{t \in I}\delta_t \geq 1,
    \end{align*}
    where $\delta_t \in \{0,1\}$ and $M \gg 0$.
    Other Boolean and temporal operators are embedded similarly, and more complex and nested STL formulas are defined recursively.

    While MIPs are NP hard, efficient branch-and-bound methods allow solving these problems to global optimality in reasonable time, provided theenvironment dynamics, predicates, and any other constraints are linear or convex. Their main drawback is the computational complexity; the number of binary variables increases with the complexity of STL specifications, finer time-discretization and trajectory length. 
    
    \subsubsection{Nonlinear programming formulation}
    An alternative to the MIP-based formulation is to use a purely continuous optimization framework, i.e. a formulation to Nonlinear Programming (NLP) problem~\citep{pant2017smooth}.
    This method replaces the non-differentiable $\min$ and $\max$ operators with their smooth approximations, enabling the use of gradient-based solvers.
   
    For example, consider again the specification $\varphi = \F_{I}(\mu(\mathbf x(t))\geq0)$. 
    A continuous non-convex sound underapproximation of robustness is as follows:
    \begin{equation*}
        \tilde{\max}_{t\in I}(\mu(\mathbf x)) = \frac{\sum_{t \in I}\mu(\mathbf x(t))\exp{\eta \mu(\mathbf x(t))}}{\sum_{t \in I} \exp{\eta \mu(\mathbf x(t))}}
    \end{equation*}
    where $\eta > 0$ controls the sharpness of the approximation. 
    This approximation is sound, guaranteeing that $\tilde{\max} ~\mathbf x \leq \max \mathbf x$.
    A smooth $\tilde{\min}$ operator is defined similarly, and nested STL formulas are defined recursively as expected.

    The main advantages of NLP-based methods are their scalability to high dimensions, potential inclusion of arbitrary differentiable constraints, and overall low computational complexity. 
    Although this method is {sound}, it is {not complete} as the resulting non-convex optimization problems only converge to local optimal solutions. 
    This formulation also require a {good} initial guess (often already satisfying $\varphi$) .

\subsubsection{Extensions and variations}

\paragraph{Feasibility and constrained optimization problem} 
In this section, we chose to focus on Problem \ref{prob:optimal_satisfaction}, however, the solution to Problems \ref{prob:feasibility} and \ref{prob:constrained_optimization} require only a minor adjustment; that is, given an STL formula $\varphi$, using $\rho_\varphi \geq 0$ as an additional constraint as opposed to maximizing $\rho_\varphi$. Similarly, since LTL and MITL specifications are in fact special cases of STL, treating them follows the same principles.

\paragraph{Scalability} 
Improving scalability of MIP-based solutions remains an active area of research. 
An interesting approach is to leverage recently proposed graphs of convex sets~\citep{marcucci2023motion}. While planning over a graph of convex sets introduces binary variables (a discrete decision on which sets to traverse), a shortest-path formulation over a discrete graph has a \emph{tight relaxation}, meaning that these problems can be solved to a high degree of accuracy by relaxing the binary variables, solving the continuous optimization problem, and simply rounding the solutions of the binary relaxation. 
Extensions to planning with temporal logic utilize the fact that product automata can be obtained from a specification and a GCS, and a satisfactory trace can be found with a shortest-path formulation~\citep{kurtz2023temporal}. Mixed-Integer programming or convex optimization can then solve these formulations. Other works have looked into how to further enable GCS with time bounds, leading to tighter MIQPs with STL~\citep{lin2024optimization}.

Another line of work towards better scalability focuses on convexification of non-convex optimization problems with spatio-temporal specifications~\citep{mao2022successive,takayama2025stlccp}. Convexified problems enjoy the solution time of convex optimizers while still being able to express nonconvexities in predicates and dynamics. However, even these methods still converge to locally optimal solutions due to local convexity.

\paragraph{Uncertainty}
Extending trajectory optimization techniques to handle uncertainty requires reasoning about probabilistic system behaviors, and hence it involves probabilistic extensions of temporal logics, such as chance constrained temporal logic (C2TL), PrSTL, or RiSTL (see Sec. \ref{sec:otherlogics}). These logics enable the evaluation of specifications over stochastic trajectories, allowing trajectory optimization to generate control inputs that maximize satisfaction under uncertainty \citep{sadigh2016safe, jha2018safeBS, matti-spatial2023}.

\paragraph{Closing the loop} This section discussed open-loop planning. Broadening our perspective, optimization-based techniques were used also for planning and control with feedback. \cite{belta2019formal} offer a comprehensive overview on the topic.

\subsection{Control certificate-based synthesis}

While previous sections focused on synthesis for planning, we now turn to synthesis of feedback control laws directly from state observations. These methods operate over short time horizons and are well-suited for low-level control within a hierarchical planning architecture, where robustness to disturbances and model uncertainty is critical. In this section, we explore how tools from nonlinear control — particularly Control Lyapunov Functions (CLFs) and Control Barrier Functions (CBFs) — can be used to enforce STL specifications in continuous-time dynamical systems. Appropriate CLFs or CBFs are constructed such that adhering to the stability and forward invariance constraints implies that the system satisfies the original specification. As such, \emph{planning} is implicit during the CLF and CBF construction.

We restrict our attention to nonlinear control-affine systems
\begin{equation*}
    \dot{x} = f(x) + g(x) u,
\end{equation*}
where $x \in \mathbb{R}^n$ is the system state, $u \in \mathbb{R}^m$ is the control input, and $f: \mathbb{R}^n \rightarrow \mathbb{R}^n$, $g: \mathbb{R}^n \rightarrow \mathbb{R}^{n \times m}$ are locally Lipschitz continuous vector fields. 
We focus on Problem \ref{prob:feasibility}, with the goal of designing a closed-loop controller $u = k(x)$ such that the closed loop system $\dot{x}=f(x) + g(x) k(x)$ satisfies a given STL specification.

\subsubsection{Control Lyapunov functions and the $\F$ operator}

CLFs are used to design feedback controllers that guarantee convergence to a desired goal state or goal set. 
Let $\mathcal{X}_g \subset \mathbb{R}^n$ be a desired goal set. A \emph{Control Lyapunov Function} $V: \mathbb{R}^n \rightarrow \mathbb{R}_{\geq 0}$ satisfies:
\begin{align*}
    V(x) \leq 0 &\iff x \in \mathcal{X}_g, \\
    V(x) > 0 &\iff x \notin \mathcal{X}_g,
\end{align*}
and for all $x \notin \mathcal{X}_g$, there exists a control input $u$ such that:
\begin{equation}
    \dot{V}(x, u) := \nabla V(x)^\top (f(x) + g(x) u) < 0.\label{eq:CLF}
\end{equation}

If such a CLF $V$ exists, then any trajectory will converge asymptotically to $\mathcal{X}_g$ \citep{khalil2002nonlinear}. 
This corresponds directly to satisfaction of STL formula $\F_{[0, \infty]} (p_\mu)$ where $p_\mu$ indicates goal satisfaction.

\subsubsection{Control barrier functions and the $\G$ operator}
CBFs are often seen as the dual to CLFs and enforce {safety} by ensuring that the system state remains within a safe set. 
Let $\mathcal{X}_s \subset \mathbb{R}^n$ be a set of safe states defined by
\begin{equation*}
    \mathcal{X}_s := \{x \in \mathbb{R}^n \mid h(x) \geq 0\},
\end{equation*}
for a continuously differentiable function $h: \mathbb{R}^n \rightarrow \mathbb{R}$. Then $h$ is a \emph{Control Barrier Function} if, for all $x \in \mathcal{X}_s$, there exists $u \in \mathbb{R}^m$ such that:
\begin{equation}
    \nabla h(x)^\top (f(x) + g(x) u) \geq -\alpha(h(x)),\label{eq:CBF}
\end{equation}
where $\alpha$ is an extended class $\mathcal{K}$ function\footnote{A function $\alpha: \mathbb{R}_{\geq 0} \rightarrow \mathbb{R}_{\geq 0}$ is said to be of class $\mathcal{K}$ if it is continuous, strictly increasing, and satisfies $\alpha(0) = 0$. Common choices include linear functions like $\alpha(s) = \gamma s$ with $\gamma > 0$.}.

\begin{definition}[Forward Invariance]
A set $\mathcal{X}_s$ is forward invariant under a control law $u(t)$ if $x(0) \in \mathcal{X}_s$ implies $x(t) \in \mathcal{X}_s$ for all $t \geq 0$.
\end{definition}

CBFs are constructed to ensure that the safe set $\mathcal{X}_s$ is forward invariant \citep{ames2016control}. 
In other words, if the system starts within $\mathcal{X}_s$, the CBF control law will keep it inside the safe set. The inequality in~\eqref{eq:CBF} guarantees that the rate of change of the barrier function $h(x)$ is nonnegative at the boundary, preventing the system from violating the safety constraint. By selecting an appropriate function $h$ and class $\mathcal{K}$ function $\alpha$, one can synthesize a feedback control law that provably enforces the safety condition. As such, CBFs serve as a natural mechanism for enforcing temporal logic specifications of the form $\G_{[0, \infty]}(p_\mu)$, where $p_\mu$ is a safety-relevant predicate (e.g., having a sufficient distance from an obstacle).

\subsubsection{Time-varying control barrier functions}
In order to synthesize control inputs in real-time, CLFs and CBFs can be combined in a quadratic program, often referred to as CLF-CBF-QP. Given some reference control input $u_{\text{ref}}$ which could be as simple as zero controls, we seek to solve the following problem:
\begin{equation}
        \begin{aligned}
        u^* = \arg  \min_{u \in \mathcal{U}}  \ \ \lVert u - u_\text{ref}\rVert_2 \quad
        \text{s.t.} \quad  \text{Eq.} \eqref{eq:CLF}, \eqref{eq:CBF}
        \end{aligned}
    \label{eq:cbf-qp}
\end{equation}
Note that the two constraints in the QP can be conflicting, meaning that it might not be possible converge to the goal set without violating safety conditions. 
A recent line of work has proposed to design a time-varying CBF that, if used for control synthesis, results in satisfaction of a complex STL formula \citep{lindemann2018control}, including conjuctions of $\G_{[a, b]} \ p_{\mu}$ and $\F_{[a, b]} \ p_{\mu}$.
In particular, a time-varying CBF takes the form of
\begin{align}
    b_p(x, t) &= \gamma(t) + \mu(x)
\end{align}
where $\gamma(t)$ is a purely time-varying function that is to be designed and $\mu(x)$ is the predicate function, i.e. $\mu(x) \geq 0 \Leftrightarrow x, t\models p_{\mu}$. The function $\gamma(t)$ should be designed so that $x \in \{x \in \mathcal{X}\mid b_p(x, t)\geq 0\} \Leftrightarrow p_{\mu}=\top$ holds either for some $t \in [a,b]$ in case of an eventually specification or for all $t\in[a,b]$ in case of always. Thus, we want the time-varying CBF to satisfy one the following conditions:
\begin{itemize}
    \item $\F_{[a, b]} p_{\mu}$: $\exists t' \in [a, b], b_1(x, t) \leq \mu(x)$
    \item $\G_{[a, b]}p_{\mu}$: $ \forall t' \in [a, b], b_1(x, t) \leq \mu(x)$.
\end{itemize}
A simple design choice for such function $\gamma(t)$ is a linear or exponential function which can be automatically constructed \citep{lindemann2020thesis}. 

Further, conjunctions of single temporal operators can be achieved by using the smooth underapproximations introduced in Sec. \ref{sec:trajopt}, to ensure continuous differentiability of the CBF. More complex nested formulas that include the until operator and nested conjunctions are also possible to handle \citep{lindemann2020thesis}.

\subsubsection{Extensions and variations}

\paragraph{Specification expressivneess}
One of the limitations of using certificate-based control synthesis is that it can only handle a restricted fragment of STL. Some works have considered LTL \citep{srinivasan2018control} or MITL \citep{8795925} specifications; some others such as \emph{Fixed-Time} and \emph{Prescribed-Time CLFs}~\citep{garg2022fixed} can ensure that convergence happens within a bounded time horizon, aligning well with bounded-time STL formulas like $\F_{[a,b]}$.
Particularly challenging in full STL are disjunctions that introduce points of non-differentiability, which cannot be directly handled by standard CBF formulations. 

To address these limitations and enlarge the admissible STL fragment, \cite{yu2024continuous} proposed the use of STL trees, which allow decomposing specifications into hierarchical structures and identifying sets that must remain invariant over prescribed time intervals. Similarly, \cite{wiltz2022handling} introduced tools from non-smooth analysis, providing a way to reason about disjunctions and capture system behavior even when classical differentiability assumptions fail.

\paragraph{Uncertainty} Another line of work focuses on extending barrier certificates for STL specifications to uncertain systems. One form of uncertainty arises from stochastic dynamics, which can be modeled using continuous-time stochastic differential equations (SDEs) 
or discrete-time stochastic difference equations.
For example, \cite{kordabad2024control} develop CBF-based methods that provide probabilistic guarantees of satisfying temporal logic constraints despite process noise and random disturbances.
Another form of uncertainty that can affect the closed-loop behavior is measurement uncertainty, which leads to noisy state estimates \citep{ruo2025robust}. 

\paragraph{CBF design} 
Finding valid CBFs for systems with input constraints is a difficult and open problem. Learning CBFs has originated for safety-critical systems \citep{dawson2023safe} but has recently been extended to temporal logic tasks \citep{liu2023learning}. Specifically, the authors directly parameterize a CBF candidate function through a neural network and train it end-to-end using BarrierNet \citep{xiao2023barriernet}.

\subsection{Main takeaways}
\begin{itemize}
\item Graph and game-based synthesis that use discrete models abstract away the nuances of dynamical models. As such, it is suitable especially for longer-horizon task-planning, high-level decision-making, and discrete motion planning from LTL.
\begin{itemize}
\item The automata-based approach relies on translating the temporal logic specification into an automaton, composing it with the discrete system model into a product automaton, and analyzing the product automaton with the use of graph search algorithms.
\item Scalability is a great challenge, especially when discrete models are more complex than deterministic transition systems. 
\end{itemize}
\item {Temporal logic goals and constraints can be integrated into sampling-based motion planning algorithms while often maintaining their complexity and completeness properties.}
\begin{itemize}
\item Quantitative semantics, like robustness of STL can be used as a criterion for rewiring, ensuring asymptotic convergence to the most satisfying path; or as a biasing or guiding criterion improving scalability. 
\item A multi-layered approach uses high-level, often automata-based synthesis together with low-level planning to ensure both specification satisfaction and geometric or kinodynamic feasibility.
\end{itemize}
\item {The main challenge of trajectory optimization with temporal logic formulas are non-convexity and non-smoothness of the optimization problem. Non-smoothness is typically addressed via encoding with binary variables or smooth non-convex approximations.}
\begin{itemize}
\item Trajectory optimization with binary variables is sound and complete under convexity assumptions, but suffers from combinatorial complexity with an increasing number of binary variables. 
\item Gradient-based optimization with smooth approximations is sound but not complete and may require good initial guesses. However, it scales well to higher dimensional systems and can consider a wide array of possible predicate and constraint functions.
\end{itemize}
\item CBFs and CLFs can be extended to obtain a feedback controller that guarantees a closed-loop system evolution to satisfy STL specifications. 
\begin{itemize}
    \item CLFs can be designed to satisfy \emph{eventually} specifications while CBFs provide a certificate for \emph{always} specifications.
    \item Both, CLFs and CBFs, can be unified into one time-varying CBF that allows satisfaction of more complex temporal specifications.
\end{itemize}
\item Many state-of-the-art solutions combine multiple of the above paradigms into hierarchical frameworks.
\end{itemize}

%% file: Sections-arxiv/extreme.tex
\newpage

\section{Uncertainty - a challenge for formal methods in real world}

Many formal synthesis techniques have shown promising results in simulation or under controlled assumptions, but their deployment in real robotic systems presents additional challenges. Strong theoretical guarantees often rely on conservative problem formulations or strong assumptions; or come at the cost of high computational demands. These limit the potential for real-world deployment. 

In this section, we discuss uncertainty as one specific and major factor that complicates the transfer of formal synthesis results. Section \ref{sec:principles} already outlined some principles of dealing with uncertainty in formal synthesis and showed it is a challenge that receives attention in all of the synthesis approaches. This section complements the previous one with discussions on how to treat different sources of uncertainty and gives examples of how formal methods are adapted, combined, and embedded towards meeting the demands of real-world robotics. 

\subsection{Uncertainty due to noisy perception}

First, uncertainty arises from imperfect perception. Maps derived from perception may be imprecise or only partially known and the true system state may not be fully observable and noise-free.
If noise is modeled as stochastic, specifications or guarantees need to be probabilistic. Extensions of temporal logic have therefore been developed to capture requirements under uncertainty — such as those that allow reasoning about the probability of a robot’s location relative to perceived landmarks in a semantic map \citep{kantaros-2022}. These formulations conceptually bridge perception and high-level reasoning, integrating semantic SLAM with formal synthesis.

A common approach to handling noisy perception is to shift from state-space to belief-space planning, where the robot reasons over a distribution of possible states. Within this framework, temporal-logic-guided planning can be performed over belief-space Markov decision processes with value iteration-based algorithms, enabling, for instance, mission planning for exploration rovers or aerial robots operating under uncertain localization \citep{nilsson2018toward}. Other formulations express requirements directly over probability distributions, for example through Gaussian Distribution Temporal Logic (GDTL), and combine sampling-based planning in belief space with local feedback control to achieve robustness to perceptual noise \citep{leahy2019controlBS}.
Trajectory planning in Gaussian belief spaces further allows reasoning about how the quality of the robot’s state estimate varies across the environment—an aspect illustrated in the following example \citep{matti-spatial2023}.

\subsubsection*{Case study: Synthesis in light-dark domain}
Light-dark domain is characterized by varying sensing conditions. In this settings, the dynamical system model is complemented by a stochastic sensor model 
describing the relation between states and sensor measurements. Using an extended Kalman Filter, we can obtain a Gaussian approximation of the robot's belief $b_t \in \mathcal{B}$, comprised of a mean state estimate and covariance matrix. 
Given an initial probability distribution over states, we can obtain the evolution of the belief over time using recursive Bayesian state estimation \cite{thrun2002probabilistic}. 

To enable reasoning over belief trajectories,  we extend Risk STL (RiSTL) to belief spaces. Specifically, we define risk predicates  over belief states and their robustness.
\begin{align}
\rho_{p_h^R}(\mathbf{b},t) = 
        \gamma - \mathcal{R}\left(-h\left(\mathbf{x}\left(t\right)\right)\right) 
\label{eq:riskpredicate}
\end{align}
where $h\left(\mathbf{x}\left(t\right)\right)$ is a predicate function with pdf $p\left(h\right)$ and $\mathcal{R}(\cdot)$ is a suitable risk measure, e.g. Value-at-Risk or Conditional Value-at-Risk, \citep{majumdar2019should}. These risk predicates can express properties like ``{the probability of the state being inside/outside region $\mathcal{O}$ is smaller than 90\%'' and can further be composed into complex temporal specifications by using the standard operators of STL.
Using robustness, RiSTL in belief space allows us to reason about how much a given specification is satisfied or violated over a finite belief trajectory $\mathbf{b} = b_t b_{t+1} \cdots b_{t+k}$.
To find a maximally robust trajectory for a specification $\varphi$ over a planning horizon of length $N$, 
we formulate and solve a trajectory optimization problem
\begin{equation}
	\begin{aligned}
	\label{eq:optproblem}
	\max_{u_{0:N-1}, \mathbf{b}_{0:N}} \quad & \rho_{\varphi} \left(\mathbf{b},0\right)\\
	\textrm{s.t.~~} \quad & b_{t+1} = \Phi\left(b_t, u_t, z_{t+1}\right) \hspace{0.5cm} &&\forall t = 0, \cdots, N-1\\
	& b_t \in \mathcal{B}, u_t \in \mathcal{U}    \hspace{0.5cm} &&\forall t = 0, \cdots, N\\
	&b_0 = b\left(t_0\right).
	\end{aligned}
\end{equation}
where $\Phi$ denotes the belief dynamics of the extended Kalman filter, $u \in \mathcal U$ is an input, $z \in Z$ is a sensor measurement.

In Fig. \ref{fig:Sims} (left), we illustrate a maximally robust trajectory that was synthesized for a risk-aware reach-avoid specification $\varphi_1$ with two sequential goals: 
\begin{align*}
     \varphi_1 = \bigwedge_{i = 1, 2} \G (\mathcal{R}(-h_{out}^{\mathcal{O}_i}(x)) \leq 0) \land
     \F (h_{in}^{\mathcal{G}_i}(x) \geq 0),
\end{align*}
with $\mathcal R  = \mathrm{VaR_{0.1}}$. 
The robot accurately localizes in light areas where the measurement noise is low. Note how maximizing the robustness of a belief RiSTL specification promotes active information gathering; this is due to the robustness of belief risk predicates growing with decreasing uncertainty. 
\begin{figure}[t]
	\scalebox{0.73}{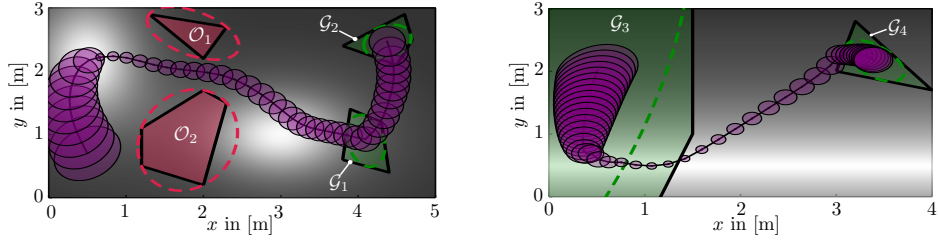}
	\caption{Light colors represent areas with low measurement noise, dark colors the opposite. The illustrated trajectories satisfy the RiSTL specifications $\varphi_1$ ({left}) and $\varphi_2$ ({right}). The 90 \% confidence ellipses about the robot's position are shown in purple, stay-in and stay-out regions are colored in green and red, respectively. The blue trajectory is shown for comparison where chance constraints (CC) are used for stay-out objectives. Adapted from \citep{matti-spatial2023}.}
	\label{fig:Sims}
\end{figure}

In another example, we require the robot to not leave area $\mathcal{G}_3$ until the state uncertainty is sufficiently low, to move cautiously (i.e. to slow down) whenever the uncertainty is above a given threshold  and to reach a goal region $\mathcal{G}_4$:
\begin{equation*}
\begin{aligned}
    \varphi_2 &= 
    (h_{in}^{\mathcal{G}_3} \geq 0\, \U \, \mathrm{tr}\left(\Sigma) \leq \varepsilon\right) \land 
    \G \left(\mathrm{tr}\left(\Sigma\right) \geq \varepsilon \Rightarrow \lVert u \rVert_2 \leq u_{max}\right) \land
    \F (h_{in}^{\mathcal{G}_4} \geq 0) ,
\end{aligned}
\end{equation*}
Here, $\mathrm{tr}(\Sigma)$ denotes the trace of the covariance matrix $\Sigma$, which serves as uncertainty quantification of the state estimate, and
$u_{max}$ is an upper bound on the controls. The optimal trajectory is illustrated
in Fig. \ref{fig:Sims} (right). Note how the robot moves slowly in the presence of large uncertainties indicated by the smaller distances between ellipses.

\subsection{Uncertainty due to external disturbances}

Another significant source of uncertainty are external disturbances. Disturbances can arise in various forms depending on the nature of the system. Examples include wind gusts, ocean currents, turbulence, slippery surfaces, etc.
Much of the research in this area work focuses on synthesizing a (maximally) robust control strategy with respect to these disturbances. 
When disturbances are modeled as stochastic processes, for example as Gaussian process noise,  optimization-based methods such as stochastic gradient descent can be used to maximize the probability that the system satisfies an STL specification \citep{scher2022robustness}. When disturbances are instead treated as bounded but unknown, reachability analysis combined with a tree-based representation of an STL formula can be employed to ensure satisfaction of the specification under the worst-case conditions. Our approach studied joint synthesis of maximal possible disturbance bounds and controllers allowing for satisfaction of STL specifications \citep{joris-disturbance2024}, discussed in more detail in the following example.

\subsubsection*{Case study: Disturbance-robust synthesis for underwater vehicle}
Autonomous Undewater Vehicles (AUVs), such as the one illustrated in Fig.~\ref{fig:Sims_underwater} are subject to underwater currents. Let us assume that this source of uncertainty is modeled as a bounded external disturbance $\mathbf d \in \mathcal{D} \subset \mathbb{R}^{n_d}$. The system can be robustly controlled with respect to an STL specification $\varphi$ if and only if
\begin{equation*}
\label{eq:NTS_satisfaction}
    \exists \mathbf u(\cdot) \in \mathcal{U}, \, \forall \mathbf d(\cdot)\in \mathcal{D}.\ \Phi(\mathbf x_0,\mathbf u(\cdot),\mathbf d(\cdot)) \models \varphi,
\end{equation*}
where $\Phi$ is a system evolution map, denoting the trajectory of the system starting from $x_0$ when applying input $\mathbf u(\cdot)$ and experiencing disturbance $\mathbf d(\cdot)$. 

\begin{figure}[t]
    \centering
    \includegraphics[width=\textwidth]{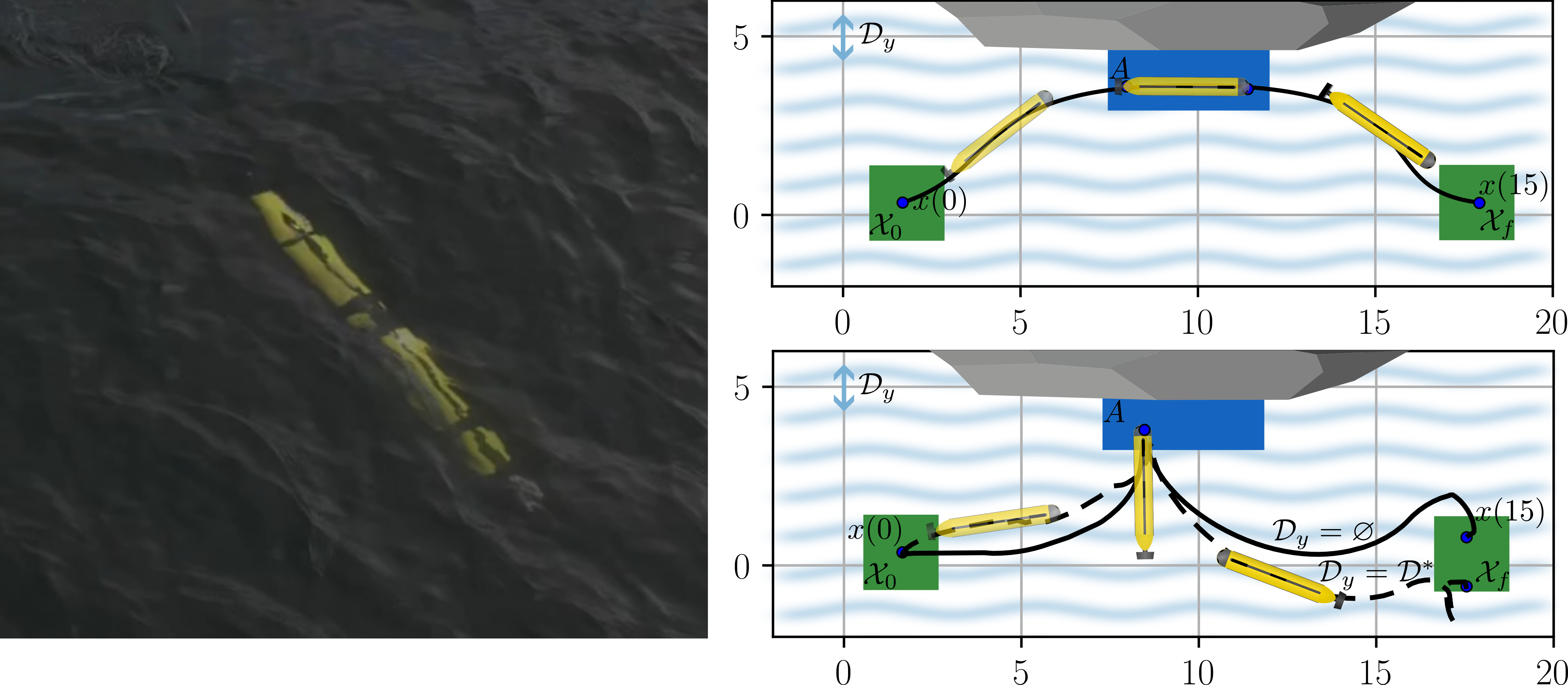}
	\caption{Small \& Affordable Maritime Underwater Robot (SAM) ({left}). A spatially robust and energy efficient trajectory ({top}) and a maximally disturbance robust trajectory ({bottom}) (without disturbances affecting the AUV as a solid line and with worst-case disturbance as a dotted line). Adapted from \citep{joris-disturbance2024}.}
	\label{fig:Sims_underwater}
\end{figure}
The disturbance set $\mathcal{D}$ may be difficult to obtain or estimate precisely enough. For example, for underwater currents this would involve finding maximal disturbances related to the waterflow and the hydrodynamics of the AUV. At the same time, conservative estimates of $\mathcal{D}$ can lead to infeasibility of any motion plan, while underapproximation of $\mathcal{D}$ might lead to catastrophic failure.

Instead of maximizing the robustness of $\varphi$, we take a different perspective and we seek for a maximal disturbance set $\hat{\mathcal{D}}$ that still allows us to satisfy $\varphi$ and a corresponding control strategy that ensures this. With this knowledge, one can make an informed decision whether to deploy an AUV in the current ocean conditions.
To solve the problem, we compute the maximal disturbance set (and the corresponding control strategy) for each reach-avoid sequence corresponding to a branch of the temporal logic tree \citep{yu2024online}, progressing in depth-first search manner.

An example of a trajectory for simple STL specification $\varphi = \G_{[7,8]} \mathbf x \in A$ is illustrated in Fig.~\ref{fig:Sims_underwater}. On top, the figure shows a trajectory that is optimal with respect to traditional STL space robustness semantics. In contrast, in the bottom the figure shows disturbance-robust behaviors. 

\subsection{Uncertainty due to modeling physical interaction}
While, on a fundamental level, any form of uncertainty can be attributed to a lack of a precise model, let us focus on the fact that physical interactions with the environment are especially difficult to model. This means that next to sufficiently robust plan, one has to pay a lot of atttention to online tracking, typically in some form of an MPC. For example, 
\cite{gu-locomotion-2025} address bipedal locomotion with the help of STL-guided trajectory optimization designed to handle translational and orientational perturbations. We addressed challenges of modeling impact dynamics, detailed in the following example \citep{joris-rss2025}.

\subsubsection*{Case study: Transportation in space via impact interactions}
Consider free-floating objects (sensor bays, toolboxes, etc.) and robots in space, each of which is given a high-level specification. For instance, a toolbox might need to be transported through the International Space Station (ISS) while avoiding unsafe regions with cluttered cables while the robot needs to ensure it has sufficient charge at all times and eventually check in on a sensor. 
In general, the object-robot specification is an STL formula
\begin{equation}
    \phi = \bigwedge_{i=1}^{N_{\textrm{obj}}}\phi_{obj_i} \land \bigwedge_{j=1}^{N_{\textrm{rob}}}\phi_{rob_j}
\end{equation}

Free-floating objects such as the toolbox do not have their own means of actuation meaning that the satisfaction of object-centric sub-specifications $\phi_{\textrm{obj}_i}$ can only be ensured via interactions -- impact with robots. We use a linear kinematic impact model $\mathcal{M}$ which may be far from the complex impact dynamics of real robots and objects. Therefore, we denote uncertainty over the post-impact velocity via an additive post-impact uncertainty term via $\dot{\mathbf x}_{obj_j}^* = \mathcal{M}(\dot{\mathbf x}_{obj_j},\dot{\mathbf x}_{rob_i}) \pm \delta$. The result is that the post-impact state is a set of possible velocities which can be propagated through the free-floating duration $\Delta t$.
Additionally, we let $\mathcal{I}$ represent whether an impact occurs between an object and robot
\begin{equation}
    \mathcal{I}(\mathbf x_{obj_j},\mathbf x_{rob_i},t) := \mathbf x_{obj_j}(t) == \mathbf x_{obj_j}(t).
\end{equation}
We then obtain conditional dynamics for the robot and objects:
\begin{align}
    \dot{\mathbf x}_{obj_j} &=\begin{cases}
        \mathcal{M}(\dot{\mathbf x}_{obj_j},\dot{\mathbf x}_{rob_i}), &\textrm{if } \mathcal{I}(\mathbf x_{obj_j},\mathbf x_{rob_i},t) \\
        0, &\textrm{otherwise}
    \end{cases} \label{eq:cond_dyn_obj}\\
    \dot{\mathbf x}_{rob_i} &=\begin{cases}
        \mathcal{M}(\dot{\mathbf x}_{rob_i},\dot{\mathbf x}_{obj_j}), &\textrm{if } \mathcal{I}(\mathbf x_{obj_j},\mathbf x_{rob_i},t) \\
        f(x,u), &\textrm{otherwise}
    \end{cases} \label{eq:cond_dyn_rob}
\end{align}
where $0$ is due to the free-flying nature of the objects in space, ensuring that objects traveling in their current direction with their current velocity remain doing so (locally).
It is via these conditional dynamics that the robots should ensure that impacts with the right object, at the right time, in the right direction with the right velocity are happening such that $\phi$ is satisfied.
This requirement leads to an incredible amount of coupled constraints with discrete decision-making that we address via the use of B\'ezier curves.

\begin{figure}[t]
    \centering
    \includegraphics[width=\textwidth]{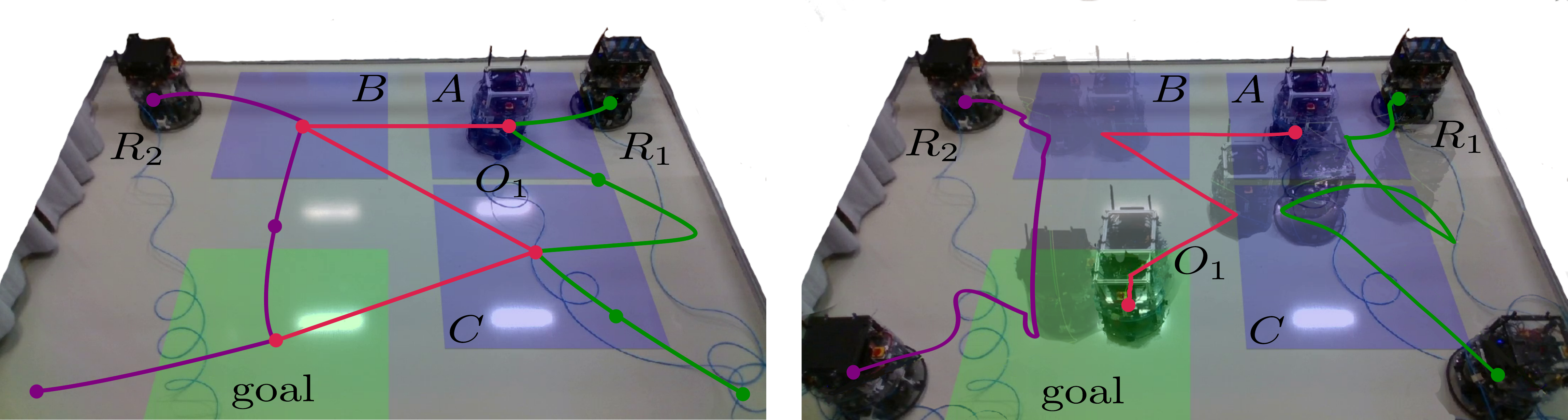}
	\caption{Preplanned and executed trajectories for the collaborative transportation task of $\phi_{obj_1}$. The endpoints of the B\'ezier trajectory segments are denoted with dots (\textbf{left}) indicating few trajectory segments for complex long-duration specifications, remedying the complexity due to its conditional dynamics.}
	\label{fig:Sims_space}
\end{figure}

We solve the following optimization problem
\begin{subequations}
\label{eq:opt_model_based}
\begin{align}
    &\max_{\mathbf x_{rob_i},\mathbf u_{rob_i},\mathbf x_{obj_j}} J, \tag{\ref*{eq:opt_model_based}} \\
    \textrm{s.t.} \quad  & \mathbf x_{rob_i} \models \phi_{rob_i} \\
    & \mathbf x_{obj_j} \models \phi_{obj_j} \\
    & \mathbf u_{rob_i} \in \mathcal{U} \\
    & \eqref{eq:cond_dyn_obj}, \eqref{eq:cond_dyn_rob}.
\end{align}
\end{subequations}
which can be encoded via Mixed-Integer Convex Programming according to \cref{sec:trajopt}.
The cost in Eq.~\eqref{eq:opt_model_based} may capture spatial robustness, temporal robustness, or the permissible additive post-impact uncertainty in the kinematic impact model, corresponding to the disturbance robustness.

Fig.~\ref{fig:Sims_space} shows a planned and executed trajectory on the ATMOS 2D space-analogue platform~\citep{roque-towards2025} where the object $O_1$ is tasked with $$\phi_{obj_1} = \bigwedge_{RoI \in[A,B,C]}\F (\mathbf x \in RoI) \land \F(\mathbf x \in \textrm{goal}).$$ 

The approximations made on the impact model, the robots trajectories, and the lack of disturbances in the offline plan from Eq.~\eqref{eq:opt_model_based} have its effect during execution on the platform, requiring online replanning and online low-level control in order to ensure the satisfaction of the specification.

%% file: Figures/MergedPlots.pdf_tex
\begingroup%
  \makeatletter%
  \providecommand\color[2][]{%
    \errmessage{(Inkscape) Color is used for the text in Inkscape, but the package 'color.sty' is not loaded}%
    \renewcommand\color[2][]{}%
  }%
  \providecommand\transparent[1]{%
    \errmessage{(Inkscape) Transparency is used (non-zero) for the text in Inkscape, but the package 'transparent.sty' is not loaded}%
    \renewcommand\transparent[1]{}%
  }%
  \providecommand\rotatebox[2]{#2}%
  \newcommand*\fsize{\dimexpr\f@size pt\relax}%
  \newcommand*\lineheight[1]{\fontsize{\fsize}{#1\fsize}\selectfont}%
  \ifx\svgwidth\undefined%
    \setlength{\unitlength}{548.02460394bp}%
    \ifx\svgscale\undefined%
      \relax%
    \else%
      \setlength{\unitlength}{\unitlength * \real{\svgscale}}%
    \fi%
  \else%
    \setlength{\unitlength}{\svgwidth}%
  \fi%
  \global\let\svgwidth\undefined%
  \global\let\svgscale\undefined%
  \makeatother%
  \begin{picture}(1,0.21140401)%
    \lineheight{1}%
    \setlength\tabcolsep{0pt}%
    \put(0,0){\includegraphics[width=\unitlength,page=1]{MergedPlots.pdf}}%
    \put(0.16328116,0.177814){\color[rgb]{1,1,1}\makebox(0,0)[lt]{\lineheight{1.25}\smash{\begin{tabular}[t]{l}$\mathcal{O}_1$\end{tabular}}}}%
    \put(0.14962554,0.08491346){\color[rgb]{1,1,1}\makebox(0,0)[lt]{\lineheight{1.25}\smash{\begin{tabular}[t]{l}$\mathcal{O}_2$\end{tabular}}}}%
    \put(0,0){\includegraphics[width=\unitlength,page=2]{MergedPlots.pdf}}%
    \put(0.29497262,0.0402789){\color[rgb]{1,1,1}\makebox(0,0)[lt]{\lineheight{1.25}\smash{\begin{tabular}[t]{l}$\mathcal{G}_1$\end{tabular}}}}%
    \put(0,0){\includegraphics[width=\unitlength,page=3]{MergedPlots.pdf}}%
    \put(0.28580286,0.18925357){\color[rgb]{1,1,1}\makebox(0,0)[lt]{\lineheight{1.25}\smash{\begin{tabular}[t]{l}$\mathcal{G}_2$\end{tabular}}}}%
    \put(0.03017783,0.01256156){\color[rgb]{0,0,0}\makebox(0,0)[lt]{\lineheight{1.25}\smash{\begin{tabular}[t]{l}0\end{tabular}}}}%
    \put(0.09922045,0.01283377){\color[rgb]{0,0,0}\makebox(0,0)[lt]{\lineheight{1.25}\smash{\begin{tabular}[t]{l}1\end{tabular}}}}%
    \put(0.17163514,0.01256156){\color[rgb]{0,0,0}\makebox(0,0)[lt]{\lineheight{1.25}\smash{\begin{tabular}[t]{l}2\end{tabular}}}}%
    \put(0.2430806,0.01256156){\color[rgb]{0,0,0}\makebox(0,0)[lt]{\lineheight{1.25}\smash{\begin{tabular}[t]{l}3\end{tabular}}}}%
    \put(0.31677077,0.01283377){\color[rgb]{0,0,0}\makebox(0,0)[lt]{\lineheight{1.25}\smash{\begin{tabular}[t]{l}4\end{tabular}}}}%
    \put(0.38944124,0.01283377){\color[rgb]{0,0,0}\makebox(0,0)[lt]{\lineheight{1.25}\smash{\begin{tabular}[t]{l}5\end{tabular}}}}%
    \put(0.18034919,0.00127006){\color[rgb]{0,0,0}\makebox(0,0)[lt]{\lineheight{1.25}\smash{\begin{tabular}[t]{l}$x$ in [m]\end{tabular}}}}%
    \put(0.00854796,0.09677812){\color[rgb]{0,0,0}\rotatebox{90.13444281}{\makebox(0,0)[lt]{\lineheight{1.25}\smash{\begin{tabular}[t]{l}$y$ in [m]\end{tabular}}}}}%
    \put(0,0){\includegraphics[width=\unitlength,page=4]{MergedPlots.pdf}}%
    \put(0.55820224,0.18878385){\color[rgb]{1,1,1}\makebox(0,0)[lt]{\lineheight{1.25}\smash{\begin{tabular}[t]{l}$\mathcal{G}_3$\end{tabular}}}}%
    \put(0,0){\includegraphics[width=\unitlength,page=5]{MergedPlots.pdf}}%
    \put(0.81810893,0.18593054){\color[rgb]{1,1,1}\makebox(0,0)[lt]{\lineheight{1.25}\smash{\begin{tabular}[t]{l}$\mathcal{G}_4$\end{tabular}}}}%
    \put(0.48630045,0.02780196){\color[rgb]{0,0,0}\makebox(0,0)[lt]{\lineheight{1.25}\smash{\begin{tabular}[t]{l}0\end{tabular}}}}%
    \put(0.48669421,0.08681429){\color[rgb]{0,0,0}\makebox(0,0)[lt]{\lineheight{1.25}\smash{\begin{tabular}[t]{l}1\end{tabular}}}}%
    \put(0.48681309,0.14595917){\color[rgb]{0,0,0}\makebox(0,0)[lt]{\lineheight{1.25}\smash{\begin{tabular}[t]{l}2\end{tabular}}}}%
    \put(0.48650852,0.20262571){\color[rgb]{0,0,0}\makebox(0,0)[lt]{\lineheight{1.25}\smash{\begin{tabular}[t]{l}3\end{tabular}}}}%
    \put(0.49631493,0.01502005){\color[rgb]{0,0,0}\makebox(0,0)[lt]{\lineheight{1.25}\smash{\begin{tabular}[t]{l}0\end{tabular}}}}%
    \put(0.58599042,0.01517598){\color[rgb]{0,0,0}\makebox(0,0)[lt]{\lineheight{1.25}\smash{\begin{tabular}[t]{l}1\end{tabular}}}}%
    \put(0.67622423,0.01502005){\color[rgb]{0,0,0}\makebox(0,0)[lt]{\lineheight{1.25}\smash{\begin{tabular}[t]{l}2\end{tabular}}}}%
    \put(0.76562729,0.01502005){\color[rgb]{0,0,0}\makebox(0,0)[lt]{\lineheight{1.25}\smash{\begin{tabular}[t]{l}3\end{tabular}}}}%
    \put(0.85490019,0.01517598){\color[rgb]{0,0,0}\makebox(0,0)[lt]{\lineheight{1.25}\smash{\begin{tabular}[t]{l}4\end{tabular}}}}%
    \put(0.65424148,0.00124374){\color[rgb]{0,0,0}\makebox(0,0)[lt]{\lineheight{1.25}\smash{\begin{tabular}[t]{l}$x$ in [m]\end{tabular}}}}%
    \put(0.47722175,0.09679273){\color[rgb]{0,0,0}\rotatebox{90.13444281}{\makebox(0,0)[lt]{\lineheight{1.25}\smash{\begin{tabular}[t]{l}$y$ in [m]\end{tabular}}}}}%
    \put(0.01837206,0.0271182){\color[rgb]{0,0,0}\makebox(0,0)[lt]{\lineheight{1.25}\smash{\begin{tabular}[t]{l}0\end{tabular}}}}%
    \put(0.01876582,0.0861305){\color[rgb]{0,0,0}\makebox(0,0)[lt]{\lineheight{1.25}\smash{\begin{tabular}[t]{l}1\end{tabular}}}}%
    \put(0.01888468,0.14527537){\color[rgb]{0,0,0}\makebox(0,0)[lt]{\lineheight{1.25}\smash{\begin{tabular}[t]{l}2\end{tabular}}}}%
    \put(0.01858011,0.20194194){\color[rgb]{0,0,0}\makebox(0,0)[lt]{\lineheight{1.25}\smash{\begin{tabular}[t]{l}3\end{tabular}}}}%
  \end{picture}%
\endgroup%

%% file: Sections-arxiv/challenges.tex
\section{Further challenges of formal synthesis in real world}

\subsection{Dynamic changes}
\label{sec:challenges-dynamic}

Robots often operate in dynamic settings -- the environment may change over time, and even if it is static, the robot’s knowledge about it may evolve. Designing synthesis methods that can deal with such changes without loosing provable guarantees remains on of the central open challenges.

Many works take the approach of adapting the plan or policy online. Techniques such as reactive planning, plan ``patching" \citep{livingston-2012-patching}, or reconfiguration \citep{guo-reconfiguration-2015} address this by repairing existing strategies rather than recomputing them from scratch. More recent works combine online learning and synthesis, enabling the robot to refine its model of the world as it operates \cite{grover2021semantic}. Other works specifically address dynamic obstacles, for example in integrated mission and motion planning frameworks \citep{alonso2018reactive} or via real-time RRT* \citep{alexis-realtimerrt2023}. 

Dynamic changes do not stem only from the environment. The robot's capabilities may change as well, for instance due to failures, requiring resilient synthesis algorithms \citep{kalluraya2023resilient}. Finally, the task specification itself may evolve over time, requiring the robot to satisfy new or modified goals, as studied in dynamic-request formulations of LTL planning \citep{vasile2020reactive}.

In general, formal synthesis in dynamic contexts faces challenges similar to those in planning and control in general -- proactive solutions may be overly inefficient both computationally and in resulting robot performance, while reactive solutions that still lead to provable specification satisfaction may not exist. An open question is how to balance these approaches toward practical, yet guarantee-preserving solutions.

\subsection{Model availability}

So far, we have focused on model-based formal synthesis, where the robot and its environment are described through analytical or symbolic models. However, in many real-world scenarios, such models are only approximate or completely unavailable. This has motivated a rapidly growing body of work on data-driven or learning-based formal synthesis, which aims to relax the dependency on explicit models while still offering provable or empirical guarantees.

Different approaches have emerged. Some expand model-based and model-free reinforcement learning to include temporal logic specification \citep{dorsa-2014-learning,aksaray-2016-QSTL} and often transform the specification into a reward structure \citep{li-2017-RL}. Others employ deep learning to learn neural netwok controllers \citep{meng2023signal,leung2023backpropagation}. 
Recent deep generative methods even embed temporal logic directly into diffusion-based policy generation \citep{ltldog}. 
Temporal logic specifications have also been integrated as a complement to learning from demonstrations \citep{wang2022cooking} as well as in LLM-driven robots \citep{yang-2024-safetychip}, to enhance the guarantees of task satisfaction or safety.

A major challenge — but also a key opportunity — lies in combining the adaptability of modern learning with the guarantees that formal synthesis offers. Learning-based approaches can handle complex, high-dimensional data and incomplete models, but they often lack provable correctness, interpretability, and reliable behavior under distributional shifts. The future of formal synthesis likely lies in bridging these strengths and weaknesses. A particular promising direction is neurosymbolic planning \citep{sun-2024-neurosymbolic}.

\subsection{Multi-robot systems}

Teams of robots can accomplish what a single robot cannot. Working together increases capabilities, efficiency, and resilience, enabling applications such as collaborative transportation, planetary exploration, and search and rescue.
The main challenge, however, is scalability: the computational burden of formal synthesis—already significant for a single robot—grows rapidly with the number of agents and their interactions. This affects all synthesis paradigms, from graph- and game-based approaches to optimization-based ones, where the joint state and specification spaces expand combinatorially.

Different strategies have been proposed to mitigate this growth. A common theme is decomposition, where the overall problem is broken into smaller subproblems that are solved separately and then composed. Broadly speaking, approaches can be characterized as top-down or bottom-up.
In top-down synthesis, a single global specification is defined for the entire team and then decomposed across robots, using methods such as automata-based decomposition \citep{chen-2012,schillinger2018simultaneous}, convex optimization \citep{charitidou2021signal} or mixed-integer programming~\citep{leahy2022fast}. 
In bottom-up approaches, each robot is given an individual specification — sometimes dependent on others — and the team behavior emerges from local guarantees or coordination rules \citep{guo2014cooperative}.

Another important distinction is between centralized and decentralized synthesis. Centralized schemes have full knowledge of all agents and can coordinate them precisely
but face severe computational and communication demands, addressed, for example, through biased sampling in sampling-based motion planning \citep{kantaros2020stylus}. Decentralized approaches, on the other hand, rely on limited communication or local information sharing, making the preservation of formal guarantees significantly more difficult. Recent efforts address this through distributed optimization and communication-aware control ~\citep{lindemann2019control,liu2020distributed}, as well as capability-based formulations that assign tasks according to the robots’ heterogeneous abilities rather than fixed identities \citep{leahy2021scalable, cardona2024planning}.

Extending formal synthesis to multi-robot systems amplifies every difficulty of the single-robot case. Coordination, communication, and heterogeneity make scalability and correctness even harder to achieve. This is why formal synthesis for multi-robot systems is a particularly active area of research.

\subsection{Human-robot interaction}

Human-Robot Interaction (HRI) represents an area where formal synthesis could have particularly strong impact: robots often operate in close proximity to humans, where safety, predictability, and trust are critical.
 Many approaches treat humans as sources of uncertainty or dynamic changes in the environment, similar as in Section \ref{sec:challenges-dynamic} and deploy reactive or adaptive planning techniques to respond to changes in real time \citep{shen-2021}. Other work explicitly models humans, either as an adversarial agent for worst-case synthesis, as a collaborative agent in a shared-autonomy settings, or as an agent with uncertain behavior \citep{fu2015synthesis, schlossman2019toward}.
Some approaches do not explicitly model the human at all but instead incorporate feedback from humans \citep{hadas-handovers} or in turn provide feedback or guidance \citep{georg-advice2023} to maintain safe and efficient interaction.

HRI presents unique opportunities for formal synthesis: from defining specifications of desired interaction to handling human adaptation and variability while ensuring robust and safe system behavior \citep{Kress2021}. 

%% file: Sections-arxiv/outro.tex
\section{Conclusions}

Formal methods have become an important tool for reasoning about and achieving complex robotic behavior with provable guarantees  ~\citep{belta2019formal,kress2018synthesis,yin-2024-review}.
Despite substantial progress, bridging the gap between formal synthesis and practical deployment remains challenging: uncertainty, model inaccuracy, and computational scalability continue to limit applicability in real-world systems. Recent work increasingly looks at how formal methods can be combined with learning, adaptation, and probabilistic modeling to address these challenges. Advancing these approaches will be essential for achieving autonomy that is  reliable and robust in practice, including in interactions with humans.

\section*{Acknowledgement}
This work was partially supported by the Wallenberg AI, Autonomous Systems and Software Program (WASP) funded by the Knut and Alice Wallenberg Foundation and by the Wenner-Gren Foundations.

%% file: references.bib
@INPROCEEDINGS{joris-disturbance2024,
  author={Verhagen, Joris and Lindemann, Lars and Tumova, Jana},
  booktitle={IEEE 63rd Conference on Decision and Control}, 
  title={Robust STL Control Synthesis under Maximal Disturbance Sets}, 
  year={2024},
  volume={},
  number={},
  pages={315-321}}

@INPROCEEDINGS{alexis-realtimerrt2023,
  author={Linard, Alexis and Torre, Ilaria and Bartoli, Ermanno and Sleat, Alex and Leite, Iolanda and Tumova, Jana},
  booktitle={IEEE/RSJ International Conference on Intelligent Robots and Systems}, 
  title={Real-Time {RRT*} with Signal Temporal Logic Preferences}, 
  year={2023},
  volume={},
  number={},
  pages={8621-8627}}

@INPROCEEDINGS{matti-spatial2023,
  author={Vahs, Matti and Pek, Christian and Tumova, Jana},
  booktitle={IEEE International Conference on Robotics and Automation}, 
  title={Risk-aware Spatio-temporal Logic Planning in Gaussian Belief Spaces}, 
  year={2023},
  volume={},
  number={},
  pages={7879-7885}}

@inproceedings{georg-advice2023,
  title={Follow my Advice: Assume-Guarantee Approach to Task Planning with Human in the Loop},
  author={Schuppe, Georg Friedrich and Torre, Ilaria and Leite, Iolanda and Tumova, Jana},
  booktitle={Robotics: Science and Systems},
  year={2023}
}

@inproceedings{joris-rss2025,
  title={Collaborative Object Transportation in Space via Impact Interactions},
  author={Verhagen, Joris and Tumova, Jana},
  booktitle={Robotics: Science and Systems},
  year={2025}
}

@INPROCEEDINGS{alexis-multiclass2022,
  author={Linard, Alexis and Torre, Ilaria and Leite, Iolanda and Tumova, Jana},
  booktitle={2022 IEEE/RSJ International Conference on Intelligent Robots and Systems}, 
  title={Inference of Multi-Class {STL} Specifications for Multi-Label Human-Robot Encounters}, 
  year={2022},
  volume={},
  number={},
  pages={1305-1311}}

@article{roque-towards2025,
  title={Towards Open-Source and Modular Space Systems with {ATMOS}},
  author={Roque, Pedro and Phodapol, Sujet and Krantz, Elias and Lim, Jaeyoung and Verhagen, Joris and Jiang, Frank J and D{\"o}rner, David and Mao, Huina and Tibert, Gunnar and Siegwart, Roland and others},
  journal={arXiv preprint arXiv:2501.16973},
  year={2025}
}

@inproceedings{schlossman2019toward,
  title={Toward achieving formal guarantees for human-aware controllers in human-robot interactions},
  author={Schlossman, Rachel and Kim, Minkyu and Topcu, Ufuk and Sentis, Luis},
  booktitle={IEEE/RSJ International Conference on Intelligent Robots and Systems},
  pages={7770--7776},
  year={2019},
  organization={IEEE}
}

@INPROCEEDINGS{hadas-handovers,
  author={Kshirsagar, Alap and Ravi, Rahul Kumar and Kress-Gazit, Hadas and Hoffman, Guy},
  booktitle={IEEE International Conference on Robot and Human Interactive Communication}, 
  title={Timing-Specified Controllers with Feedback for Human-Robot Handovers}, 
  year={2022},
  volume={},
  number={},
  pages={1313-1320}}

@ARTICLE{chen-2012,
  author={Chen, Yushan and Ding, Xu Chu and Stefanescu, Alin and Belta, Calin},
  journal={IEEE Transactions on Robotics}, 
  title={Formal Approach to the Deployment of Distributed Robotic Teams}, 
  year={2012},
  volume={28},
  number={1},
  pages={158-171}}

@ARTICLE{sun-2024-neurosymbolic,
  author={Sun, Xiaowu and Shoukry, Yasser},
  journal={IEEE Transactions on Robotics}, 
  title={Neurosymbolic Motion and Task Planning for Linear Temporal Logic Tasks}, 
  year={2024},
  volume={40},
  number={},
  pages={2749-2768}}

@INPROCEEDINGS{yang-2024-safetychip,
  author={Yang, Ziyi and Raman, Shreyas S. and Shah, Ankit and Tellex, Stefanie},
  booktitle={IEEE International Conference on Robotics and Automation}, 
  title={Plug in the Safety Chip: Enforcing Constraints for {LLM}-driven Robot Agents}, 
  year={2024},
  volume={},
  number={},
  pages={14435-14442}}

@inproceedings{li-2017-RL, 
author = {Li, Xiao and Vasile, Cristian-Ioan and Belta, Calin}, title = {Reinforcement learning with temporal logic rewards}, year = {2017}, booktitle = {IEEE/RSJ International Conference on Intelligent Robots and Systems}, pages = {3834–3839}, numpages = {6}, location = {Vancouver, BC, Canada} }

@INPROCEEDINGS{dorsa-2014-learning,
  author={Sadigh, Dorsa and Kim, Eric S. and Coogan, Samuel and Sastry, S. Shankar and Seshia, Sanjit A.},
  booktitle={53rd IEEE Conference on Decision and Control}, 
  title={A learning based approach to control synthesis of Markov decision processes for linear temporal logic specifications}, 
  year={2014},
  volume={},
  number={},
  pages={1091-1096}}

@INPROCEEDINGS{aksaray-2016-QSTL,
  author={Aksaray, Derya and Jones, Austin and Kong, Zhaodan and Schwager, Mac and Belta, Calin},
  booktitle={IEEE Conference on Decision and Control}, 
  title={Q-Learning for robust satisfaction of signal temporal logic specifications}, 
  year={2016},
  volume={},
  number={},
  pages={6565-6570}
}

@ARTICLE{ltldog,
  author={Feng, Zeyu and Luan, Hao and Goyal, Pranav and Soh, Harold},
  journal={IEEE Robotics and Automation Letters}, 
  title={{LTLDoG}: Satisfying Temporally-Extended Symbolic Constraints for Safe Diffusion-Based Planning}, 
  year={2024},
  volume={9},
  number={10},
  pages={8571-8578}}

@article{guo-reconfiguration-2015,
author = {Meng Guo and Dimos V Dimarogonas},
title ={Multi-agent plan reconfiguration under local LTL specifications},

journal = {The International Journal of Robotics Research},
volume = {34},
number = {2},
pages = {218-235},
year = {2015}
}

@INPROCEEDINGS{livingston-2012-patching,
  author={Livingston, Scott C. and Murray, Richard M. and Burdick, Joel W.},
  booktitle={IEEE International Conference on Robotics and Automation}, 
  title={Backtracking temporal logic synthesis for uncertain environments}, 
  year={2012},
  volume={},
  number={},
  pages={5163-5170}}

@inproceedings{grover2021semantic,
  title={Semantic abstraction-guided motion planningfor {scLTL} missions in unknown environments},
  author={Grover, Kush and Barbosa, Fernando S and Tumova, Jana and Kret{\i}nsky, Jan},
  booktitle={Robotics: Science and Systems},
  year={2021}
}

@article{bombara21, author = {Bombara, Giuseppe and Belta, Calin}, title = {Offline and Online Learning of Signal Temporal Logic Formulae Using Decision Trees}, year = {2021}, issue_date = {July 2021}, publisher = {Association for Computing Machinery}, address = {New York, NY, USA}, volume = {5}, number = {3}, journal = {ACM Transactions on Cyber-Physical Systems}, month = mar, articleno = {22}, numpages = {23} }

@InProceedings{aasi-2023,
  title = 	 {Time-Incremental Learning of Temporal Logic Classifiers Using Decision Trees},
  author =       {Aasi, Erfan and Cai, Mingyu and Vasile, Cristian Ioan and Belta, Calin},
  booktitle = 	 {Annual Learning for Dynamics and Control Conference},
  pages = 	 {547--559},
  year = 	 {2023},
  volume = 	 {211},
  series = 	 {Proceedings of Machine Learning Research},
  publisher =    {PMLR}
}

@ARTICLE{gu-locomotion-2025,
  author={Gu, Zhaoyuan and Zhao, Yuntian and Chen, Yipu and Guo, Rongming and Leestma, Jennifer K. and Sawicki, Gregory S. and Zhao, Ye},
  journal={IEEE Transactions on Robotics}, 
  title={Robust-Locomotion-By-Logic: Perturbation-Resilient Bipedal Locomotion via Signal Temporal Logic Guided Model Predictive Control}, 
  year={2025},
  volume={41},
  number={},
  pages={4300-4321}}

@INPROCEEDINGS{lindemann-2020acc,
  author={Lindemann, Lars and Dimarogonas, Dimos V},
  booktitle={2020 American Control Conference}, 
  title={Efficient Automata-based Planning and Control under Spatio-Temporal Logic Specifications}, 
  year={2020},
  volume={},
  number={},
  pages={4707-4714}}

@INPROCEEDINGS{ho-2022cdc,
  author={Ho, Qi Heng and Ilyes, Roland B. and Sunberg, Zachary N. and Lahijanian, Morteza},
  booktitle={IEEE 61st Conference on Decision and Control}, 
  title={Automaton-Guided Control Synthesis for Signal Temporal Logic Specifications}, 
  year={2022},
  volume={},
  number={},
  pages={3243-3249}}

@ARTICLE{gundana-2021stl,
  author={Gundana, David and Kress-Gazit, Hadas},
  journal={IEEE Robotics and Automation Letters}, 
  title={Event-Based Signal Temporal Logic Synthesis for Single and Multi-Robot Tasks}, 
  year={2021},
  volume={6},
  number={2},
  pages={3687-3694}}

@INPROCEEDINGS{liu-2014mtl,
  author={Liu, Jun and Prabhakar, Pavithra},
  booktitle={IEEE International Conference on Robotics and Automation}, 
  title={Switching control of dynamical systems from metric temporal logic specifications}, 
  year={2014},
  volume={},
  number={},
  pages={5333-5338}}

@inproceedings{bouyer-2006mtl,
  author       = {Patricia Bouyer and
                  Laura Bozzelli and
                  Fabrice Chevalier},
  title        = {Controller Synthesis for {MTL} Specifications},
  booktitle    = {International Conference on Concurrency Theory},
  series       = {Lecture Notes in Computer Science},
  volume       = {4137},
  pages        = {450--464},
  publisher    = {Springer},
  year         = {2006}
}

@ARTICLE{pacheck-2023-repair,
  author={Pacheck, Adam and Kress-Gazit, Hadas},
  journal={IEEE Transactions on Robotics}, 
  title={Physically Feasible Repair of Reactive, Linear Temporal Logic-Based, High-Level Tasks}, 
  year={2023},
  volume={39},
  number={6},
  pages={4653-4670},
  doi={10.1109/TRO.2023.3304009}}

@INPROCEEDINGS{li-2011-mining,
  author={Li, Wenchao and Dworkin, Lili and Seshia, Sanjit A.},
  booktitle={ACM/IEEE International Conference on Formal Methods and Models for Codesign}, 
  title={Mining assumptions for synthesis}, 
  year={2011},
  volume={},
  number={},
  pages={43-50}}

@INPROCEEDINGS{fainekos-2011revising,
  author={Fainekos, Georgios E.},
  booktitle={2011 IEEE International Conference on Robotics and Automation}, 
  title={Revising temporal logic specifications for motion planning}, 
  year={2011},
  volume={},
  number={},
  pages={40-45}}

@article{yin-2024-review,
title = {Formal synthesis of controllers for safety-critical autonomous systems: Developments and challenges},
journal = {Annual Reviews in Control},
volume = {57},
pages = {100940},
year = {2024},
author = {Xiang Yin and Bingzhao Gao and Xiao Yu}
}

@inproceedings{nilsson2018toward,
  title={Toward Specification-Guided Active Mars Exploration for Cooperative Robot Teams.},
  author={Nilsson, Petter and Haesaert, Sofie and Thakker, Rohan and Otsu, Kyohei and Vasile, Cristian Ioan and Agha-Mohammadi, Ali-Akbar and Murray, Richard M and Ames, Aaron D},
  booktitle={Robotics: Science and systems},
  volume={14},
  pages={1--9},
  year={2018}
}

@ARTICLE{luo-2021-abstractionfree,
  author={Luo, Xusheng and Kantaros, Yiannis and Zavlanos, Michael M.},
  journal={IEEE Transactions on Robotics}, 
  title={An Abstraction-Free Method for Multirobot Temporal Logic Optimal Control Synthesis}, 
  year={2021},
  volume={37},
  number={5},
  pages={1487-1507}}

@ARTICLE{kantaros-2022,
  author={Kantaros, Yiannis and Kalluraya, Samarth and Jin, Qi and Pappas, George J.},
  journal={IEEE Transactions on Robotics}, 
  title={Perception-Based Temporal Logic Planning in Uncertain Semantic Maps}, 
  year={2022},
  volume={38},
  number={4},
  pages={2536-2556}}

@article{alonso2018reactive,
  title={Reactive mission and motion planning with deadlock resolution avoiding dynamic obstacles},
  author={Alonso-Mora, Javier and DeCastro, Jonathan A and Raman, Vasumathi and Rus, Daniela and Kress-Gazit, Hadas},
  journal={Autonomous Robots},
  volume={42},
  number={4},
  pages={801--824},
  year={2018},
  publisher={Springer}
}

@INPROCEEDINGS{shen-2021,
  author={Li, Shen and Park, Daehyung and Sung, Yoonchang and Shah, Julie A. and Roy, Nicholas},
  booktitle={IEEE International Conference on Robotics and Automation}, 
  title={Reactive Task and Motion Planning under Temporal Logic Specifications}, 
  year={2021},
  volume={},
  number={},
  pages={12618-12624}}

@InProceedings{plaku-2012-LTL,
author="Plaku, Erion",
title="Planning in Discrete and Continuous Spaces: From LTL Tasks to Robot Motions",
booktitle="Advances in Autonomous Robotics",
year="2012",
publisher="Springer Berlin Heidelberg",
address="Berlin, Heidelberg",
pages="331--342"
}

@misc{wang2025conformalnl2ltltranslatingnaturallanguage,
      title={{ConformalNL2LTL}: Translating Natural Language Instructions into Temporal Logic Formulas with Conformal Correctness Guarantees}, 
      author={Jun Wang and David Smith Sundarsingh and Jyotirmoy V. Deshmukh and Yiannis Kantaros},
      year={2025},
      eprint={2504.21022},
      archivePrefix={arXiv},
      primaryClass={cs.CL},
      url={https://arxiv.org/abs/2504.21022}, 
}

@inproceedings{ltllfd-shah2018,
 author = {Shah, Ankit and Kamath, Pritish and Shah, Julie A and Li, Shen},
 booktitle = {Advances in Neural Information Processing Systems},
 pages = {},
 publisher = {Curran Associates, Inc.},
 title = {Bayesian Inference of Temporal Task Specifications from Demonstrations},

 volume = {31},
 year = {2018}
}

@INPROCEEDINGS{ltlviz,
  author={Srinivas, Shashank and Kermani, Ramtin and Kim, Kangjin and Kobayashi, Yoshihiro and Fainekos, Georgios},
  booktitle={IEEE International Conference on Robotics and Biomimetics}, 
  title={A graphical language for {LTL} motion and mission planning}, 
  year={2013},
  volume={},
  number={},
  pages={704-709}}

@InProceedings{robustness,
author="Donz{\'e}, Alexandre
and Maler, Oded",
editor="Chatterjee, Krishnendu
and Henzinger, Thomas A.",
title="Robust Satisfaction of Temporal Logic over Real-Valued Signals",
booktitle="Formal Modeling and Analysis of Timed Systems",
year="2010",
publisher="Springer Berlin Heidelberg",
address="Berlin, Heidelberg",
pages="92--106",
isbn="978-3-642-15297-9"
}

@ARTICLE{nok-rh2012,
  author={Wongpiromsarn, Tichakorn and Topcu, Ufuk and Murray, Richard M.},
  journal={IEEE Transactions on Automatic Control}, 
  title={Receding Horizon Temporal Logic Planning}, 
  year={2012},
  volume={57},
  number={11},
  pages={2817-2830}}

@ARTICLE{ltlmdp,
  author={Ding, Xuchu and Smith, Stephen L. and Belta, Calin and Rus, Daniela},
  journal={IEEE Transactions on Automatic Control}, 
  title={Optimal Control of {Markov} Decision Processes With Linear Temporal Logic Constraints}, 
  year={2014},
  volume={59},
  number={5},
  pages={1244-1257}}

@article{chou2022learning,
  title={Learning temporal logic formulas from suboptimal demonstrations: theory and experiments},
  author={Chou, Glen and Ozay, Necmiye and Berenson, Dmitry},
  journal={Autonomous Robots},
  volume={46},
  number={1},
  pages={149--174},
  year={2022},
  publisher={Springer}
}

@article{cristi-twt2017,
title = {Time window temporal logic},
journal = {Theoretical Computer Science},
volume = {691},
pages = {27-54},
year = {2017},
author = {Cristian-Ioan Vasile and Derya Aksaray and Calin Belta}
}

@ARTICLE{noushin-wstl2021,
  author={Mehdipour, Noushin and Vasile, Cristian-Ioan and Belta, Calin},
  journal={IEEE Control Systems Letters}, 
  title={Specifying User Preferences Using Weighted Signal Temporal Logic}, 
  year={2021},
  volume={5},
  number={6},
  pages={2006-2011}}

@ARTICLE{rodionova-temporalSTL2023,
  author={Rodionova, Alëna and Lindemann, Lars and Morari, Manfred and Pappas, George J.},
  journal={IEEE Control Systems Letters}, 
  title={Combined Left and Right Temporal Robustness for Control Under STL Specifications}, 
  year={2023},
  volume={7},
  number={},
  pages={619-624}}

@INPROCEEDINGS{nawaz-interactive2024,
  author={Nawaz, Farhad and Peng, Shaoting and Lindemann, Lars and Figueroa, Nadia and Matni, Nikolai},
  booktitle={IEEE/RSJ International Conference on Intelligent Robots and Systems}, 
  title={Reactive Temporal Logic-based Planning and Control for Interactive Robotic Tasks}, 
  year={2024},
  volume={},
  number={},
  pages={12108-12115}}

@inproceedings{
wang2022cooking,
title={Temporal Logic Imitation: Learning Plan-Satisficing Motion Policies from Demonstrations},
author={Yanwei Wang and Nadia Figueroa and Shen Li and Ankit Shah and Julie Shah},
booktitle={6th Annual Conference on Robot Learning},
year={2022}
}

@INPROCEEDINGS{vasilopoulos-manipulation2021,
  author={Vasilopoulos, Vasileios and Kantaros, Yiannis and Pappas, George J. and Koditschek, Daniel E.},
  booktitle={IEEE International Conference on Robotics and Automation}, 
  title={Reactive Planning for Mobile Manipulation Tasks in Unexplored Semantic Environments}, 
  year={2021},
  volume={},
  number={},
  pages={6385-6392}}

@ARTICLE{morteza-pctl2012,
  author={Lahijanian, Morteza and Andersson, Sean B. and Belta, Calin},
  journal={IEEE Transactions on Robotics}, 
  title={Temporal Logic Motion Planning and Control With Probabilistic Satisfaction Guarantees}, 
  year={2012},
  volume={28},
  number={2},
  pages={396-409}}

@InProceedings{maler-monitoring,
author="Maler, Oded
and Nickovic, Dejan",
title="Monitoring Temporal Properties of Continuous Signals",
booktitle="Formal Techniques, Modelling and Analysis of Timed and Fault-Tolerant Systems",
year="2004",
publisher="Springer Berlin Heidelberg",
address="Berlin, Heidelberg",
pages="152--166"
}

@ARTICLE{hadas-tro09,
  author={Kress-Gazit, Hadas and Fainekos, Georgios E. and Pappas, George J.},
  journal={IEEE Transactions on Robotics}, 
  title={Temporal-Logic-Based Reactive Mission and Motion Planning}, 
  year={2009},
  volume={25},
  number={6},
  pages={1370-1381}}

@article{steve-ijrr2011, 
author = {Stephen L Smith and Jana Tůmová and Calin Belta and Daniela Rus},
title ={Optimal path planning for surveillance with temporal-logic constraints},
journal = {The International Journal of Robotics Research},
volume = {30},
number = {14},
pages = {1695-1708},
year = {2011},
}

@inproceedings{maly-2013,
author = {Maly, Matthew R. and Lahijanian, Morteza and Kavraki, Lydia E. and Kress-Gazit, Hadas and Vardi, Moshe Y.},
title = {Iterative temporal motion planning for hybrid systems in partially unknown environments},
year = {2013},
publisher = {Association for Computing Machinery},
address = {New York, NY, USA},
booktitle = {Proceedings of the 16th International Conference on Hybrid Systems: Computation and Control},
pages = {353–362},
numpages = {10},
location = {Philadelphia, Pennsylvania, USA}
}

@INPROCEEDINGS{cristi-samplingmaxsat2017,
  author={Vasile, Cristian-Ioan and Raman, Vasumathi and Karaman, Sertac},
  booktitle={IEEE/RSJ International Conference on Intelligent Robots and Systems}, 
  title={Sampling-based synthesis of maximally-satisfying controllers for temporal logic specifications}, 
  year={2017},
  volume={},
  number={},
  pages={3840-3847}}

@ARTICLE{Kantaros-samplingmultiTRO2019,
  author={Kantaros, Yiannis and Zavlanos, Michael M.},
  journal={IEEE Transactions on Automatic Control}, 
  title={Sampling-Based Optimal Control Synthesis for Multirobot Systems Under Global Temporal Tasks}, 
  year={2019},
  volume={64},
  number={5},
  pages={1916-1931},
  doi={10.1109/TAC.2018.2853558}}

@INPROCEEDINGS{bhatia-sampling2010,
  author={Bhatia, Amit and Kavraki, Lydia E. and Vardi, Moshe Y.},
  booktitle={IEEE International Conference on Robotics and Automation}, 
  title={Sampling-based motion planning with temporal goals}, 
  year={2010},
  volume={},
  number={},
  pages={2689-2696}}

@INPROCEEDINGS{plaku-sampling2014,
  author={McMahon, James and Plaku, Erion},
  booktitle={ IEEE/RSJ International Conference on Intelligent Robots and Systems}, 
  title={Sampling-based tree search with discrete abstractions for motion planning with dynamics and temporal logic}, 
  year={2014},
  volume={},
  number={},
  pages={3726-3733}}

@article{ZHANG2020105591,
title = {Randomized sampling-based trajectory optimization for UAVs to satisfy linear temporal logic specifications},
journal = {Aerospace Science and Technology},
volume = {96},
pages = {105591},
year = {2020},
author = {Zetian Zhang and Ruixiang Du and Raghvendra V. Cowlagi}
}

@ARTICLE{kyunghoon-cost2017,
  author={Cho, Kyunghoon and Suh, Junghun and Tomlin, Claire J. and Oh, Songhwai},
  journal={IEEE Robotics and Automation Letters}, 
  title={Cost-Aware Path Planning Under Co-Safe Temporal Logic Specifications}, 
  year={2017},
  volume={2},
  number={4},
  pages={2308-2315}}

@INPROCEEDINGS{cristi-sampling2013,
  author={Vasile, Cristian-Ioan and Belta, Calin},
  booktitle={2013 IEEE/RSJ International Conference on Intelligent Robots and Systems}, 
  title={Sampling-based temporal logic path planning}, 
  year={2013},
  volume={},
  number={},
  pages={4817-4822}}

@article{marchesini2025sampling,
  title={Sampling-Based Planning Under {STL} Specifications: A Forward Invariance Approach},
  author={Marchesini, Gregorio and Liu, Siyuan and Lindemann, Lars and Dimarogonas, Dimos V},
  journal={arXiv preprint arXiv:2506.10739},
  year={2025}
}

@INPROCEEDINGS{luis-leastviolating203,
  author={Reyes Castro, Luis I. and Chaudhari, Pratik and Tůmová, Jana and Karaman, Sertac and Frazzoli, Emilio and Rus, Daniela},
  booktitle={52nd IEEE Conference on Decision and Control}, 
  title={Incremental sampling-based algorithm for minimum-violation motion planning}, 
  year={2013},
  volume={},
  number={},
  pages={3217-3224}}

@inproceedings{chatrola2025kinodynamic, 
author = {Chatrola, Jeel and Ajith, Abhiroop and Leahy, Kevin and Chamzas, Constantinos}, title = {Multi-layer Motion Planning with Kinodynamic and Spatio-Temporal Constraints}, year = {2025}, publisher = {Association for Computing Machinery}, booktitle = {ACM International Conference on Hybrid Systems: Computation and Control}, articleno = {5}, numpages = {7}}

@book{baier2008principles,
	title        = {Principles of model checking},
	author       = {Baier, Christel and Katoen, Joost-Pieter},
	year         = 2008,
	publisher    = {MIT press}
}

@inproceedings{guo2014cooperative,
	title        = {Cooperative decentralized multi-agent control under local {LTL} tasks and connectivity constraints},
	author       = {Guo, Meng and Tumova, Jana and Dimarogonas, Dimos V},
	year         = 2014,
	booktitle    = {53rd IEEE Conference on Decision and Control},
	pages        = {75--80},
	organization = {IEEE}
}

@inproceedings{kloetzer2008dealing,
	title        = {Dealing with nondeterminism in symbolic control},
	author       = {Kloetzer, Marius and Belta, Calin},
	year         = 2008,
	booktitle    = {International Workshop on Hybrid Systems: Computation and Control},
	pages        = {287--300},
	organization = {Springer}
}

@article{kress2018synthesis,
  title={Synthesis for robots: Guarantees and feedback for robot behavior},
  author={Kress-Gazit, Hadas and Lahijanian, Morteza and Raman, Vasumathi},
  journal={Annual Review of Control, Robotics, and Autonomous Systems},
  volume={1},
  pages={211--236},
  year={2018},
  publisher={Annual Reviews}
}

@article{kantaros2020stylus,
	title        = {Stylus*: A temporal logic optimal control synthesis algorithm for large-scale multi-robot systems},
	author       = {Kantaros, Yiannis and Zavlanos, Michael M},
	year         = 2020,
	journal      = {The International Journal of Robotics Research},
	publisher    = {SAGE Publications Sage UK: London, England},
	volume       = 39,
	number       = 7,
	pages        = {812--836}
}

@inproceedings{donze2010robust,
	title        = {Robust satisfaction of temporal logic over real-valued signals},
	author       = {Donz{\'e}, Alexandre and Maler, Oded},
	year         = 2010,
	booktitle    = {Proceedings of the International Conference on Formal Modeling and Analysis of Timed Systems},
	pages        = {92--106}
}

@inproceedings{finucane2010ltlmop,
	title        = {{LTLMoP}: Experimenting with language, temporal logic and robot control},
	author       = {Finucane, Cameron and Jing, Gangyuan and Kress-Gazit, Hadas},
	year         = 2010,
	booktitle    = {Proceedings of the IEEE/RSJ International Conference on Intelligent Robots and Systems},
	pages        = {1988--1993}
}

@article{lindemann2018control,
	title        = {Control barrier functions for signal temporal logic tasks},
	author       = {Lindemann, Lars and Dimarogonas, Dimos V},
	year         = 2018,
	journal      = {IEEE control systems letters},
	volume       = 3,
	number       = 1,
	pages        = {96--101}
}

@article{Kress2021,
	title        = {Formalizing and guaranteeing human-robot interaction},
	author       = {Kress-Gazit, Hadas and Eder, Kerstin and Hoffman, Guy and Admoni, Henny and Argall, Brenna and Ehlers, Ruediger and Heckman, Christoffer and Jansen, Nils and Knepper, Ross and K{\v{r}}et{\'\i}nský, Jan and others},
	year         = 2021,
	journal      = {Communications of the ACM},
	volume       = 64,
	number       = 9,
	pages        = {78--84}
}

@article{vasile2020reactive,
	title        = {Reactive sampling-based path planning with temporal logic specifications},
	author       = {Vasile, Cristian Ioan and Li, Xiao and Belta, Calin},
	year         = 2020,
	journal      = {The International Journal of Robotics Research},
	publisher    = {SAGE Publications Sage UK: London, England},
	volume       = 39,
	number       = 8,
	pages        = {1002--1028}
}

@inproceedings{tumova2013least,
  title={Least-violating control strategy synthesis with safety rules},
  author={Tumova, Jana and Hall, Gavin C and Karaman, Sertac and Frazzoli, Emilio and Rus, Daniela},
  booktitle={Proceedings of the 16th international conference on Hybrid systems: computation and control},
  pages={1--10},
  year={2013}
}

@inproceedings{sadigh2016safe,
  title={Safe control under uncertainty with probabilistic signal temporal logic},
  author={Sadigh, Dorsa and Kapoor, Ashish},
  booktitle={Proceedings of Robotics: Science and Systems XII},
  year={2016}
}

@article{charitidou2021signal,
  title={Signal temporal logic task decomposition via convex optimization},
  author={Charitidou, Maria and Dimarogonas, Dimos V},
  journal={IEEE Control Systems Letters},
  volume={6},
  pages={1238--1243},
  year={2021},
  publisher={IEEE}
}

@article{leahy2022fast,
  title={Fast decomposition of temporal logic specifications for heterogeneous teams},
  author={Leahy, Kevin and Jones, Austin and Vasile, Cristian-Ioan},
  journal={IEEE Robotics and Automation Letters},
  volume={7},
  number={2},
  pages={2297--2304},
  year={2022},
  publisher={IEEE}
}

@article{lindemann2019control,
  title={Control barrier functions for multi-agent systems under conflicting local signal temporal logic tasks},
  author={Lindemann, Lars and Dimarogonas, Dimos V},
  journal={IEEE control systems letters},
  volume={3},
  number={3},
  pages={757--762},
  year={2019},
  publisher={IEEE}
}

@article{belta2019formal,
  title={Formal methods for control synthesis: An optimization perspective},
  author={Belta, Calin and Sadraddini, Sadra},
  journal={Annual Review of Control, Robotics, and Autonomous Systems},
  volume={2},
  number={1},
  pages={115--140},
  year={2019},
  publisher={Annual Reviews}
}

@article{liu2020distributed,
  title={Distributed communication-aware motion planning for networked mobile robots under formal specifications},
  author={Liu, Zhiyu and Wu, Bo and Dai, Jin and Lin, Hai},
  journal={IEEE Transactions on Control of Network Systems},
  volume={7},
  number={4},
  pages={1801--1811},
  year={2020},
  publisher={IEEE}
}

@article{garg2022fixed,
  title={Fixed-time control under spatiotemporal and input constraints: A quadratic programming based approach},
  author={Garg, Kunal and Arabi, Ehsan and Panagou, Dimitra},
  journal={Automatica},
  volume={141},
  pages={110314},
  year={2022},
  publisher={Elsevier}
}

@article{meng2023signal,
  title={Signal temporal logic neural predictive control},
  author={Meng, Yue and Fan, Chuchu},
  journal={IEEE Robotics and Automation Letters},
  volume={8},
  number={11},
  pages={7719--7726},
  year={2023},
  publisher={IEEE}
}

@article{cardona2024planning,
  title={Planning for heterogeneous teams of robots with temporal logic, capability, and resource constraints},
  author={Cardona, Gustavo A and Vasile, Cristian-Ioan},
  journal={The International Journal of Robotics Research},
  volume={43},
  number={13},
  pages={2089--2111},
  year={2024},
  publisher={SAGE Publications Sage UK: London, England}
}

@inproceedings{chen2018signal,
  title={Signal temporal logic meets reachability: Connections and applications},
  author={Chen, Mo and Tam, Qizhan and Livingston, Scott C and Pavone, Marco},
  booktitle={International Workshop on the Algorithmic Foundations of Robotics},
  pages={581--601},
  year={2018},
  organization={Springer}
}

@article{yu2024online,
  title={Online control synthesis for uncertain systems under signal temporal logic specifications},
  author={Yu, Pian and Gao, Yulong and Jiang, Frank J and Johansson, Karl H and Dimarogonas, Dimos V},
  journal={The International Journal of Robotics Research},
  volume={43},
  number={6},
  pages={765--790},
  year={2024},
  publisher={SAGE Publications Sage UK: London, England}
}

@inproceedings{raman2014model,
  title={Model predictive control from signal temporal logic specifications: A case study},
  author={Raman, Vasumathi and Maasoumy, Mehdi and Donz{\'e}, Alexandre},
  booktitle={Proceedings of the 4th ACM SIGBED international workshop on design, modeling, and evaluation of cyber-physical systems},
  pages={52--55},
  year={2014}
}

@inproceedings{pant2017smooth,
  title={Smooth operator: Control using the smooth robustness of temporal logic},
  author={Pant, Yash Vardhan and Abbas, Houssam and Mangharam, Rahul},
  booktitle={2017 IEEE Conference on Control Technology and Applications (CCTA)},
  pages={1235--1240},
  year={2017},
  organization={IEEE}
}

@inproceedings{mao2022successive,
  title={Successive convexification for optimal control with signal temporal logic specifications},
  author={Mao, Yuanqi and Acikmese, Behcet and Garoche, Pierre-Loic and Chapoutot, Alexandre},
  booktitle={Proceedings of the 25th ACM International Conference on Hybrid Systems: Computation and Control},
  pages={1--7},
  year={2022}
}

@article{takayama2025stlccp,
  author={Takayama, Yoshinari and Hashimoto, Kazumune and Ohtsuka, Toshiyuki},
  journal={IEEE Transactions on Automatic Control}, 
  title={{STLCCP:} Efficient Convex Optimization-Based Framework for Signal Temporal Logic Specifications}, 
  year={2025},
  volume={70},
  number={9},
  pages={6064-6079}}

@article{leung2023backpropagation,
  title={Backpropagation through signal temporal logic specifications: Infusing logical structure into gradient-based methods},
  author={Leung, Karen and Ar{\'e}chiga, Nikos and Pavone, Marco},
  journal={The International Journal of Robotics Research},
  volume={42},
  number={6},
  pages={356--370},
  year={2023},
  publisher={SAGE Publications Sage UK: London, England}
}

@inproceedings{srinivasan2018control,
  title={Control of multi-agent systems with finite time control barrier certificates and temporal logic},
  author={Srinivasan, Mohit and Coogan, Samuel and Egerstedt, Magnus},
  booktitle={2018 IEEE Conference on Decision and Control},
  pages={1991--1996},
  year={2018},
  organization={IEEE}
}

@INPROCEEDINGS{8795925,
  author={Barbosa, Fernando S. and Lindemann, Lars and Dimarogonas, Dimos V. and Tumova, Jana},
  booktitle={European Control Conference}, 
  title={Integrated Motion Planning and Control Under Metric Interval Temporal Logic Specifications}, 
  year={2019},
  volume={},
  number={},
  pages={2042-2049}}

@article{leahy2021scalable,
  title={Scalable and robust algorithms for task-based coordination from high-level specifications (Scratches)},
  author={Leahy, Kevin and Serlin, Zachary and Vasile, Cristian-Ioan and Schoer, Andrew and Jones, Austin M and Tron, Roberto and Belta, Calin},
  journal={IEEE Transactions on Robotics},
  volume={38},
  number={4},
  pages={2516--2535},
  year={2021},
  publisher={IEEE}
}

@article{schillinger2018simultaneous,
  title={Simultaneous task allocation and planning for temporal logic goals in heterogeneous multi-robot systems},
  author={Schillinger, Philipp and B{\"u}rger, Mathias and Dimarogonas, Dimos V},
  journal={The international journal of robotics research},
  volume={37},
  number={7},
  pages={818--838},
  year={2018},
  publisher={Sage Publications Sage UK: London, England}
}

@article{marcucci2023motion,
  title={Motion planning around obstacles with convex optimization},
  author={Marcucci, Tobia and Petersen, Mark and von Wrangel, David and Tedrake, Russ},
  journal={Science robotics},
  volume={8},
  number={84},
  pages={eadf7843},
  year={2023},
  publisher={American Association for the Advancement of Science}
}

@article{kurtz2023temporal,
  title={Temporal logic motion planning with convex optimization via graphs of convex sets},
  author={Kurtz, Vince and Lin, Hai},
  journal={IEEE Transactions on Robotics},
  volume={39},
  number={5},
  pages={3791--3804},
  year={2023},
  publisher={IEEE}
}

@inproceedings{lin2024optimization,
  author={Lin, Xuan and Ren, Jiming and Coogan, Samuel and Zhao, Ye},
  booktitle={2025 IEEE International Conference on Robotics and Automation}, 
  title={Optimization-Based Task and Motion Planning Under Signal Temporal Logic Specifications Using Logic Network Flow}, 
  year={2025},
  volume={},
  number={},
  pages={12565-12571},
}

@article{leahy2019controlBS,
  title={Control in belief space with temporal logic specifications using vision-based localization},
  author={Leahy, Kevin and Cristofalo, Eric and Vasile, Cristian-Ioan and Jones, Austin and Montijano, Eduardo and Schwager, Mac and Belta, Calin},
  journal={The International Journal of Robotics Research},
  volume={38},
  number={6},
  pages={702--722},
  year={2019},
  publisher={SAGE Publications Sage UK: London, England}
}

@ARTICLE{9293004,
  author={Oh, Yoonseon and Cho, Kyunghoon and Choi, Yunho and Oh, Songhwai},
  journal={IEEE Transactions on Automatic Control}, 
  title={Chance-Constrained Multilayered Sampling-Based Path Planning for Temporal Logic-Based Missions}, 
  year={2021},
  volume={66},
  number={12},
  pages={5816-5829},
  doi={10.1109/TAC.2020.3044273}}

@article{jha2018safeBS,
  title={Safe autonomy under perception uncertainty using chance-constrained temporal logic},
  author={Jha, Susmit and Raman, Vasumathi and Sadigh, Dorsa and Seshia, Sanjit A},
  journal={Journal of Automated Reasoning},
  volume={60},
  number={1},
  pages={43--62},
  year={2018},
  publisher={Springer}
}

@inproceedings{chatterjee2015qualitative,
  title={Qualitative analysis of {POMDP}s with temporal logic specifications for robotics applications},
  author={Chatterjee, Krishnendu and Chmelik, Martin and Gupta, Raghav and Kanodia, Ayush},
  booktitle={IEEE International Conference on Robotics and Automation},
  pages={325--330},
  year={2015},
  organization={IEEE}
}

@inproceedings{ho2023planning,
  title={Planning with SiMBA: Motion Planning under Uncertainty for Temporal Goals using Simplified Belief Guides},
  author={Ho, Qi Heng and Sunberg, Zachary N and Lahijanian, Morteza},
  booktitle={2023 IEEE International Conference on Robotics and Automation},
  pages={5723--5729},
  year={2023},
  organization={IEEE}
}

@article{thrun2002probabilistic,
  title={Probabilistic robotics},
  author={Thrun, Sebastian},
  journal={Communications of the ACM},
  volume={45},
  number={3},
  pages={52--57},
  year={2002},
  publisher={ACM New York, NY, USA}
}

@article{safaoui2020control,
  title={Control design for risk-based signal temporal logic specifications},
  author={Safaoui, Sleiman and Lindemann, Lars and Dimarogonas, Dimos V and Shames, Iman and Summers, Tyler H},
  journal={IEEE Control Systems Letters},
  volume={4},
  number={4},
  pages={1000--1005},
  year={2020},
  publisher={IEEE}
}

@inproceedings{majumdar2019should,
author="Majumdar, Anirudha
and Pavone, Marco",
title="How Should a Robot Assess Risk? {T}owards an Axiomatic Theory of Risk in Robotics",
booktitle="Robotics Research",
year="2020",
publisher="Springer International Publishing",
address="Cham",
pages="75--84",
}

@book{khalil2002nonlinear,
  title={Nonlinear systems},
  author={Khalil, Hassan K and Grizzle, Jessy W},
  volume={3},
  year={2002},
  publisher={Prentice hall Upper Saddle River, NJ}
}

@article{ames2016control,
  title={Control barrier function based quadratic programs for safety critical systems},
  author={Ames, Aaron D and Xu, Xiangru and Grizzle, Jessy W and Tabuada, Paulo},
  journal={IEEE Transactions on Automatic Control},
  volume={62},
  number={8},
  pages={3861--3876},
  year={2016},
  publisher={IEEE}
}

@phdthesis{lindemann2020thesis,
  title={Planning and control of multi-agent systems under signal temporal logic specifications},
  author={Lindemann, Lars},
  year={2020},
  school={KTH Royal Institute of Technology}
}

@inproceedings{liu2023learning,
  title={Learning robust and correct controllers from signal temporal logic specifications using barriernet},
  author={Liu, Wenliang and Xiao, Wei and Belta, Calin},
  booktitle={IEEE Conference on Decision and Control},
  pages={7049--7054},
  year={2023},
  organization={IEEE}
}

@article{yu2024continuous,
  title={Continuous-time control synthesis under nested signal temporal logic specifications},
  author={Yu, Pian and Tan, Xiao and Dimarogonas, Dimos V},
  journal={IEEE Transactions on Robotics},
  volume={40},
  pages={2272--2286},
  year={2024},
  publisher={IEEE}
}

@inproceedings{wiltz2022handling,
  title={Handling disjunctions in signal temporal logic based control through nonsmooth barrier functions},
  author={Wiltz, Adrian and Dimarogonas, Dimos V},
  booktitle={IEEE Conference on Decision and Control},
  pages={3237--3242},
  year={2022},
  organization={IEEE}
}

@article{xiao2023barriernet,
  title={Barriernet: Differentiable control barrier functions for learning of safe robot control},
  author={Xiao, Wei and Wang, Tsun-Hsuan and Hasani, Ramin and Chahine, Makram and Amini, Alexander and Li, Xiao and Rus, Daniela},
  journal={IEEE Transactions on Robotics},
  volume={39},
  number={3},
  pages={2289--2307},
  year={2023},
  publisher={IEEE}
}

@article{dawson2023safe,
  title={Safe control with learned certificates: A survey of neural {L}yapunov, barrier, and contraction methods for robotics and control},
  author={Dawson, Charles and Gao, Sicun and Fan, Chuchu},
  journal={IEEE Transactions on Robotics},
  volume={39},
  number={3},
  pages={1749--1767},
  year={2023},
  publisher={IEEE}
}

@inproceedings{kordabad2024control,
  title={Control barrier functions for stochastic systems under signal temporal logic tasks},
  author={Kordabad, Arash Bahari and Charitidou, Maria and Dimarogonas, Dimos V and Soudjani, Sadegh},
  booktitle={2024 European Control Conference},
  pages={3213--3219},
  year={2024},
  organization={IEEE}
}

@inproceedings{ruo2025robust,
  title={Robust {CBF}-based {STL} motion planning for socially responsible robot navigation in the presence of measurement noise},
  author={Ruo, Andrea and Sabattini, Lorenzo},
  booktitle={American Control Conference (ACC)},
  pages={402--407},
  year={2025},
  organization={IEEE}
}

@inproceedings{scher2022robustness,
  title={Robustness-based synthesis for stochastic systems under signal temporal logic tasks},
  author={Scher, Guy and Sadraddini, Sadra and Kress-Gazit, Hadas},
  booktitle={2022 IEEE/RSJ International Conference on Intelligent Robots and Systems (IROS)},
  pages={1269--1275},
  year={2022},
  organization={IEEE}
}

@article{fu2015synthesis,
  title={Synthesis of shared autonomy policies with temporal logic specifications},
  author={Fu, Jie and Topcu, Ufuk},
  journal={IEEE Transactions on Automation Science and Engineering},
  volume={13},
  number={1},
  pages={7--17},
  year={2015},
  publisher={IEEE}
}

@inproceedings{kalluraya2023resilient,
  title={Resilient temporal logic planning in the presence of robot failures},
  author={Kalluraya, Samarth and Pappas, George J and Kantaros, Yiannis},
  booktitle={IEEE Conference on Decision and Control},
  pages={7520--7526},
  year={2023},
  organization={IEEE}
}
